\documentclass{ieeetmlcn}
\usepackage{cite}
\usepackage{amsmath,amssymb,amsfonts}
\usepackage{algorithmic}
\usepackage{graphicx,color}
\usepackage{soul}
\usepackage{textcomp}
\usepackage{xcolor}
\usepackage{hyperref}
\hypersetup{hidelinks=true}
\usepackage{algorithm,algorithmic}
\def\BibTeX{{\rm B\kern-.05em{\sc i\kern-.025em b}\kern-.08em
    T\kern-.1667em\lower.7ex\hbox{E}\kern-.125emX}}
\AtBeginDocument{\definecolor{tmlcncolor}{cmyk}{0.93,0.59,0.15,0.02}\definecolor{NavyBlue}{RGB}{0,86,125}}

\usepackage{comment}

\usepackage{mdframed}

\newcommand{\newtext}[1]{\textcolor{blue}{#1}}

\newtheorem{remarks}{Remark}%[section]

\def\OJlogo{\vspace{-4pt}\includegraphics[height=18pt]{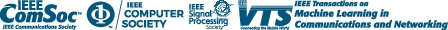}}
\def\seclogo{\vspace{10pt}\includegraphics[height=18pt]{new_logo}}

\def\authorrefmark#1{\ensuremath{^{\textbf{#1}}}}

\begin{document}
\receiveddate{XX Month, XXXX}
\reviseddate{XX Month, XXXX}
\accepteddate{XX Month, XXXX}
\publisheddate{XX Month, XXXX}
\currentdate{XX Month, XXXX}
\doiinfo{TMLCN.2022.1234567}

\markboth{}{Author {et al.}}

\title{Peering Partner Recommendation for ISPs using Machine Learning
}

\author{Md Ibrahim Ibne Alam\authorrefmark{1},
Ankur Senapati\authorrefmark{2}, Member, IEEE, 
\\ Anindo Mahmood\authorrefmark{3},
 Murat Yuksel\authorrefmark{3}, Senior Member, IEEE, \\
and Koushik Kar\authorrefmark{1}, Senior Member, IEEE}
\affil{Electrical Engineering, and Computer, and Systems Engineering, Rensselaer Polytechnic Institute, Troy, NY 12180, USA}
\affil{School of Electrical and Computer Engineering, Purdue University, West Lafayette, IN 47907, USA }
\affil{Department of Electrical and Computer Engineering, University of Central Florida, Orlando, FL 32816, USA}
\corresp{Corresponding author: Md Ibrahim Ibne Alam (email: amitu91191@gmail.com).}
% \authornote{This paragraph of the first footnote will contain support information, including sponsor and financial support acknowledgment. For example, ``This work was supported in part by the U.S. Department of Commerce under Grant 123456.''}

\begin{abstract}
% These instructions give you guidelines for preparing papers for IEEE Transactions on Machine Learning in Communications and Networking.
%  Otherwise, use this document as an instruction set. The electronic file of your paper will be formatted further at IEEE. Paper titles should be written in uppercase and lowercase letters, not all uppercase. Avoid writing long formulas with subscripts in the title; short formulas that identify the elements are fine (e.g., ``Nd--Fe--B''). Do not write ``(Invited)'' in the title. Full names of authors are preferred in the author field, but are not required. Put a space between authors' initials. The abstract must be a concise yet comprehensive reflection of what is in your article. 
Internet service providers (ISPs) need to connect with other ISPs to provide global connectivity services to their users. To ensure global connectivity, ISPs can either use transit service(s) or establish direct peering relationships between themselves via Internet exchange points (IXPs). Peering offers more room for ISP-specific optimizations and is preferred, but it often involves a lengthy and complex process. Automating peering partner selection can enhance efficiency in the global Internet ecosystem. We explore the use of publicly available data on ISPs to develop a machine learning (ML) model that can predict whether an ISP pair should peer or not. 
At first, we explore public databases, e.g., PeeringDB, CAIDA, etc., to gather data on ISPs. Then, we evaluate the performance of three broad types of ML models for predicting peering relationships: tree-based, neural network-based, and transformer-based. Among these, we observe that tree-based models achieve the highest accuracy and efficiency in our experiments. 
The XGBoost model trained with publicly available data showed promising performance, with a 98\% accuracy rate in predicting peering partners. In addition, the model demonstrated great resilience to variations in time, space, and missing data. 
We envision that ISPs can adopt our method to fully automate the peering partner selection process, thus transitioning to a more efficient and optimized Internet ecosystem.
% In particular, the abstract must be self-contained, without abbreviations, footnotes, or
% references. It should be a microcosm of the full article. The abstract must be between 150–250 words.
% Be sure that you adhere to these limits; otherwise, you will need to edit your abstract accordingly. The
% abstract must be written as one paragraph
 %, and should not contain displayed mathematical equations or tabular material. The abstract should include three or four different keywords or phrases, as this will help readers to find it. It is important to avoid over-repetition of such phrases as this can result in a page being rejected by search engines. Ensure that your abstract reads well and is grammatically correct.
\end{abstract}

\begin{IEEEkeywords}
Internet exchange points, Internet peering, Internet service provider.
\end{IEEEkeywords}


\maketitle

\section{INTRODUCTION}
% \IEEEPARstart{I}{nternet} connectivity has been the backbone that revolutionized all sectors of life. The whole world is now connected through the Internet, and hundreds of companies may incur huge losses even if there is a millisecond delay in the Internet. Different protocols and services have been proposed to facilitate smoother Internet connection and mitigate disruption or delay in the Internet. Internet Service Providers (ISPs) are companies that help end users connect to the Internet, and historically, transit providers have helped ISPs to connect. Over time, ISPs have flourished, and instead of depending on Transit Providers to carry their traffic, they are trying to be independent. To formulate that, ISPs are now peering with each other at different facilities that are usually controlled by Internet Exchange Points (IXPs). These IXPs do not carry the traffic but rather rent the facilities and switches where the ISPs can come and exchange their traffic. 

\IEEEPARstart{T}{he} modern Internet landscape is characterized by an exponential growth of data traffic \cite{meekerinternet}. Facilitated by the widespread usage of connected Internet of Things (IoT) devices, demand for online streaming services, and the rise of cloud computing platforms, the substantial increase in traffic load has put enormous strains on the existing infrastructure. There is no single Internet Service Provider (ISP) that is capable of managing extensive network backbones individually to support this increased data demand \cite{metz2001interconnecting}. Most providers utilize strategic interconnection agreements to ensure global connectivity to their customer base. Two key processes that facilitate interconnection between ISPs are: i) peering agreement and ii) transit agreement \cite{norton2001internet, myIX_forum}. 

The idea of peering was initially implemented by ISPs around the beginning of this millennium to save on transit costs \cite{norton2001internet}. At that time, larger ISPs that provided transit services charged smaller ISPs for carrying their traffic and enabling global Internet connectivity. For worldwide connectivity, Tier 1 transit ISPs were inter-connected via peering (one-to-one connection), which was predominant within these types of ISPs. With the increasing popularity of content delivery networks (CDNs), content ISPs preferred the more profitable peering connection with Access ISPs (also known as Eyeballs). During the last two decades, the hierarchical (vertical) architecture of ISPs has evolved to a more horizontal architecture where a large portion of the ISPs are peering with each other \cite{dey2019peering}. More formally, `peering' between two or more ISPs is a legal contract between the ISPs to exchange their traffic directly (through switches or routers) without the involvement/help of any other ISPs.
%(usually the transit providers). 
Currently, while ISPs use both peering and transit options, peering is becoming increasingly important in the Internet ecosystem \cite{loye2022global, prehn2022peering}.

In the 2000s, the cost of sending traffic through transit services was almost 10 times higher than peering \cite{norton2001internet}. That ratio fell to 2.5--1 around 2015 and nowadays the per-unit transit cost is usually less than the peering costs \cite{Hurr_Elec, IXP_Mbps_cost, FDC}. However, even with lower monetary costs, ISPs are more inclined towards peering. There are quite a few advantages of peering over transit services, and different studies, some conducted by Internet eXchange Providers (IXPs), have identified the following advantages of peering \cite{thousand_eye_isp_peering, ams_ix_why}: i) resilient interconnection and better reachability, ii)  increased routing control resulting in reducing latency, iii) direct control over inter-traffic flow, iv) reduced cost when sharing bulk traffic in a pair, and v) secured traffic.

Due to the advantages of peering and its economic significance, a faster and automated peering process is the demand of time \cite{alam2024meta}. According to \cite{norton2001internet}, the peering process consists of three stages: ``i) Identification (Traffic Engineering Data Collection and Analysis), ii) Contact \& Qualification (Initial Peering Negotiation), and iii) Implementation Discussion (Peering Methodology). The first phases identify the {\em who} and the {\em why}, while the last phase focuses on the {\em how}." In a more recent forum on peering strategies, it was pointed out that the peering decision process can vary depending on the perspective of the ISP, i.e., if the ISP is Tier 1 and wants to peer with another Tier 1, or the ISP is a smaller one and is trying to peer with a larger ISP \cite{myIX_forum}. The characteristics or features of each ISP are different from each other, which leads to the complexity of developing a global peering process that is suitable for all ISP pairs. In general, ISPs go through a long and expensive process to find suitable peering partners and locations. To solve this complicated problem, we explore the feasibility of using publicly available data on ISPs to generate a universal model that quantifies the suitability of two ISPs for peering\footnote{\newtext{Automation reduces labor costs and improves scalability, as demonstrated by BGP, which enables automated, scalable routing across the Internet \cite{alfroy2024next}. However, this comes at the cost of increased vulnerability to attacks like BGP hijacking and route leaks. Despite these risks, BGP is widely used due to its scalability benefits. Similarly, automating ISP peering may introduce security challenges, but it has the potential to significantly scale beyond the limits of current manual practices.}}.

% At the beginning ISPs were happy just with the price reduction. However with the increasing popularity of content delivery networks (CDNs), the economy largely shifted and CDNs started making large profits compared to access provider ISPs. 

% What is peering, defining peering and peering locations (PoPs). 

% Why peering is important and how automation of peering can help the internet eco-system.

% How peering has evolved over the years. The hierarchical architecture to flatten..

% \as{An ASN acts as a unique identifier for every ISP in the network market. Their purpose relies solely on allowing Autonomous Systems (AS) to exchange routing information with other systems.}

% \vspace{10pt}
% \noindent\textbf{\textit{Available Resources:}}
% \vspace{-7pt}
% \begin{itemize}
%     \item There are large repositories of publicly available data on the ASNs (e.g., PeeringDB, CAIDA), which can be used as characterization features of the ASNs.
%     \item There are a few work on the prediction of peering relationship between two ASNs using heuristic approaches using PeeringDB data \cite{alam2024meta, dey2021meta}.
%     \item There is another work \cite{mustafa2022peer} that looked into the same problem from a machine learning perspective using data from PeeringDB and CAIDA.
%     \item There is scope of exploration on the peering prediction problem from the perspective of ML.
% \end{itemize}

Several decades of Internet measurement studies have resulted in publicly available large data repositories on Autonomous Systems (ASes), e.g., PeeringDB, CAIDA, RouteViews, and RIPE RIS. Given the abundance of public data available on ASes, we aim to answer the question: ``\textit{To what extent can the peering relationship between two ISPs be predicted?}'' We study and extract important features of ASes from these data repositories and explore \newtext{well established} machine learning methods to make the most accurate predictions on whether or not an AS pair should peer. Our framework can assist ISPs in identifying suitable peering partners. Our preliminary work \cite{mustafa2022peer} on the use of ML for peering partner prediction showed very promising results. In this work, we extend the concept to larger feature sets and systematically build a framework to yield ML models for peering partner prediction. Our main contributions include:
\begin{itemize}
       \item Investigate different ISP databases for authentic feature extraction to use in the ML-based peering prediction.
    \item %The previous work \cite{mustafa2022peer} on the peering prediction problem with ML focused on their method with processed data from PeeringDB and CAIDA, 
    Using the PeeringDB and CAIDA datasets, we start the ML modeling effort with raw features directly extracted from those two databases and systematically evaluate the impact of processing data on the performance of the ML models.
    \item Explore several processed datasets, including the one in \cite{mustafa2022peer}, and determine the best dataset for the peering partner prediction problem.
    \item Study the importance of features collected from the datasets and build a correlation matrix to identify the importance of groups of features in peering partner prediction.
    \item Compare the performance of five different ML-based models that fall under three broad ML model types.
    % , i.e., tree-based, neural network-based, and transformer-based.
    % \item Show that CAIDA features are very important and mainly focusing on those features are good enough for peering prediction.
    % \item The addition of balanced accuracy, peering and non-peering accuracy with the error bar for different data size (not explored in previous work) gives us more insight on the data dependency.
    % \item Transfer learning, the most important contribution, is used to check if ML models can predict peering prediction when one of the ASN or both ASNs in the ASN pair is unknown; and shows promising results.
    % \item As an Ablation study we checked the effect of ISP types, oversampling and undersampling, and correlation between ASN number and ASN size.
    % \item Study the change in feature characteristics (of the ASNs, collected from the databases) over time.
    \item Study transfer learning with the trained ML model and check immunity of the model over time and data.
    \item Compare the performance of our final ML model with the state-of-the-art heuristic and ML-based approaches \cite{alam2024meta, dey2021meta, mustafa2022peer}. With our final model, we achieved an overall accuracy of 98\% in predicting peering partners, which is significantly better than the methods previously proposed by others.
\end{itemize}

\newtext{Overall, the main novelty of our work is to demonstrate that there are publicly available features of ISPs that can be leveraged intelligently to attain a {\em very high} peering partner prediction accuracy while maintaining a minimal feature set. Moreover, the robustness study of our model provides great insights, including the feasibility and effectiveness of transfer learning across different temporal datasets.}
% \hl{The rest of the paper is organized as follows: ... }
The rest of the paper is organized as follows: In Section \ref{sec:related-work}, we look into existing works from the literature that aim to solve the peering partner identification problem. In Section \ref{sec:data}, we study multiple publicly available datasets that provide insight into ISPs' operational and business practices and perform a deeper analysis of the CAIDA and PeeringDB datasets to extract appropriate features. In Section \ref{sec:ablation-study}, we perform an ablation study to obtain the optimum feature set and test it on multiple machine learning algorithms. In Section \ref{sec:robustness}, we evaluate the robustness of our approach against training data size, transfer learning, time variance of the training dataset and missing data points. Finally, Section \ref{sec:conclusion} concludes the paper.
\section{Related Work}
\label{sec:related-work}
% \iia{Anindo, may be you did not get time to add related works, I added a few, please try to add more}

% For any ISP, finding appropriate peering partner(s) is crucial to survive in the market. 
% The idea of whom to peer, why, and how has been studied extensively over the last two decades. From the beginning of the peering concept, Norton has repeatedly published extensive studies on peering practices, and most of today’s work (on peering) is indebted to his works [Norton 2001, 2010]. ISPs and IXPs invest in research for peering-related topics, i.e., peering partner selection, peering policies, peering to transit comparison, etc. Workshops are arranged to know the current thoughts of the ISPs, providers, and end users. The procedure of how an ISP can identify potential peer and peering locations is also discussed in those workshops [MyIX].

The traditional peering approach is time-consuming and involves a lot of manual input. ISPs are required to go through in-person negotiations to formulate peering agreements. Once agreed, setting up peering sessions also requires manual effort from network engineers. As the size and complexity of the Internet continues to grow, this approach is becoming highly inefficient, and automating the entire process is the next logical step. The automation will need to be carried out in two fundamental aspects: 1) finding optimal peering partners and locations without any in-person contact, and 2) setting up and monitoring peering sessions autonomously. Large players of the industry have already invested heavily on solving the second step. Examples include: \emph{Meta's} session automation framework, Peering Manager (utilized by large IXPs like DE-CIX and LINX), and Noction’s Intelligent Routing Platform (IRP) \cite{2021-jenny-facebook-automation, 2021-peering-manager, 2018-ethan-nocton-irp}. However, the peer selection problem has received little attention. The work in \cite{2004-amini-peering-multi-provider} uses server capacity of providers, network delay, and traffic distribution to determine potential peers. The issue with this model is that it utilizes a relatively simple linear prediction model to estimate future traffic based on historical patterns, which can produce incorrect results during instances of traffic surges. The works in \cite{hsu2008isp, chang2006peer,  ma2008interconnecting, nomikos2018peer} improves on that but generally focus on either specific peering models (i.e, remote), peering between certain provider types (e.g, access-content) or when peering is confined to a local area. 

% \iia{needs paraphrasing} \textcolor{cyan}{Since one of the key motives behind peering is to reduce cost, there has been extensive research on game-theoretic modeling of peering ~\cite{secci2010peering}, and understanding the economics behind pricing where multiple ISPs are involved ~\cite{2006-shakkottai-economics, 
% alam2023pricing}. 
% However, the topic of automated peer selection on a global scale using (publicly) available data has not received enough attention. The few works that address issues broadly related to automated peer selection only consider some specific types of peering (i.e., remote peering), certain ISP types (access-content), or are confined to a local area \cite{hsu2008isp, chang2006peer,  ma2008interconnecting, nomikos2018peer}.}

The work of \cite{dey2021meta} is the foremost in exploring the process of peering automation on a global scale. It introduced the term meta-peering as the automation of all four phases of the peering process: the pre-peering phase, peer selection phase, the establishment of BGP
% \textcolor{blue} {Border Gateway Protocol} (BGP) 
sessions, and post-peering phase. The model introduces a heuristic algorithm that quantifies the willingness to peer between two providers and the potential increase in coverage through peering to determine the likelihood of peering. In \cite{alam2022modeling}, another heuristic approach was proposed that utilized a novel metric called stability, in addition to peering willingness, to measure the feasibility of peering between ISPs. These models were further improved in \cite{alam2024meta} through a combined heuristic-machine learning approach. The new framework fine-tuned the predictions from the heuristics with the help of machine learning algorithms to enhance peering partner prediction accuracy. 

% As a recent and much-extended work on meta-peering, the efficacy of two meta-peering methods is studied in [TNSM]. The results showed great potential in automating the selection of peering partners and good peering locations. 

\newtext{In the works of \cite{alam2024meta, alam2022modeling}, the domain knowledge was used to build heuristics to make the peering decisions. For instance, the size similarity and overlap in coverage proved to be very effective in predicting peering decisions. However, in this study, we take an ``objective'' approach in comprehensively evaluating the parameters/features involved.}
Moreover, a key limitation of these heuristic approaches are that they do not consider \newtext{some important features like the size of customers, peers, and peering policies of providers in the decision process, which can be attributed to their relatively modest prediction accuracies. However, datasets like PeeringDB lists some of the peering policies (not all), and CAIDA provides information on the number of customer and peers.} Consequently, standalone machine learning approaches that utilize these publicly available datasets about ISPs and IXPs offer a promising alternative for predicting peers. Our prior work \cite{mustafa2022peer} investigated such an approach on a limited extent and, to our knowledge, is the only one that tried to implement such a model on a global scale with publicly available data. In this work, we greatly extend the idea introduced in \cite{mustafa2022peer} and add more novel approaches.
% Our prior work [Shahzeb] investigated the automation of the peering process from the perspective of machine learning and our current work is mostly inspired by that work. The ML-based procedure followed in [] was novel and we could not find any other study that tried implementing such a model on a global scale with publicly available data. However, in this work, we greatly extend the idea introduced in [] and add more novel approaches.
% to show the potential of using publicly available data with machine learning. 

First, we perform an extensive study on the ISP features publicly available from PeeringDB and CAIDA, followed by an ablation study to assess the significance of these features in the peer selection. Second, we conduct a performance comparison of five different types of ML models and identify XGBoost (XGB) as the most suitable model for the peering prediction task. Third, from our study of ISP features, we intelligently choose the optimal dataset that ensures close to optimum accuracy while using limited features. Finally, we perform a robustness study to show the immunity of the trained ML model over time and data. \newtext{The results from this analysis is very significant in the rapidly evolving Internet landscape and was missing in \cite{mustafa2022peer}.} Moreover, the work in \cite{mustafa2022peer} focused on 15,000 ISP pairs, whereas we focused on a much larger set of around 37,000 ISP pairs. With the XGB model, we were able to attain a notable 98\% accuracy in the peering partner selection, which significantly outperforms the earlier work (reported at around 89\% accuracy).

\section{Data for Peering Partner Prediction}
\label{sec:data}

\begin{figure}[t]
    \centering
    \includegraphics[width=\linewidth]{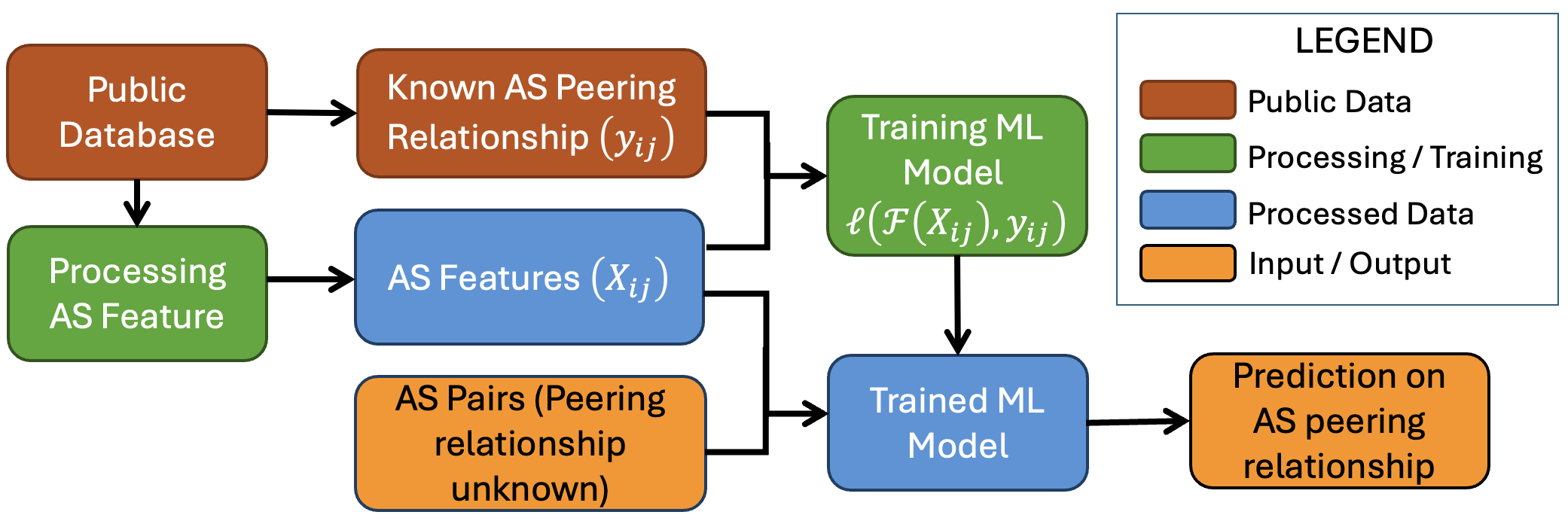}
    \caption{Framework of the proposed methodology.}
    \label{fig:framework}
\end{figure}

\subsection{Problem Formulation} \label{subsec:problem_form}
An Internet Service Provider (ISP) consists of one or more Autonomous Systems (ASes). These ASes operate independently and have specific characteristics and functional objectives. In real life, instead of ISPs peering with another ISP, the ASes pair up to peer\footnote{Larger ISPs can own \emph{different types} of Autonomous Systems (ASes) that have corresponding (unique) Autonomous System Numbers (ASNs). Hence, it is not straightforward to tag an ISP with any specific service. On the other hand, ASes (and corresponding ASNs) always follow a single role, and we can get the ASN type for most ASNs from PeeringDB \cite{PeeringDB} (some of the ASNs, especially from developing countries, are not registered in PeeringDB). Furthermore, peering policies are ultimately implemented among ASes even though the contracts are made among ISPs. Due to these reasons, we used ASNs instead of an ISP or the entire organization.}. The peering decision between two ASes is a pair-wise decision and both ASes need to agree to form the peering connection. Looking from the perspective of one AS: the AS usually performs rigorous calculations of cost and utility to check if peering with some other AS is beneficial or not. If the attributes of the ASes are known, it may be possible to come up with a generalized way to analyze those attributes and predict whether the corresponding AS pair should peer or not. We have found quite a few databases where the attributes or features of different ASes are publicly available (more discussion of this is given in Section \ref{subsec:dataset}). In addition, the peering relationship between different ASes is available from the {\em CAIDA AS relationship} dataset \cite{caida-as-rel}. We propose to use the attributes of ASes from different databases along with the AS pair peering status data (from the CAIDA AS relationship data) to train an ML-based model to predict if two ASes should peer or not. 
%More precisely, for the AS pairs, we will use the attributes (features) of the AS pairs as the input feature for the ML model along with the current peering relationships of those ASes (from CAIDA AS relationship data). 
Upon enough training, we expect the ML model to perform well on unseen data and ultimately help us automate the peering decision process.

Let us denote the feature set of AS $i$ as $f_i$. Then for two ASes $i$ and $j$, we denote $X_{ij} =f_i \cup f_j$ %\{f_i, f_j\}$ (merging of $f_i$ and $f_j$) 
as the feature set of the AS pair $(i,j)$ . For each of the $(i,j)$ AS pairs, the peering relationship information informs us if the AS pair is peering or not. Let us denote that peering status as $y_{ij}$. We propose to train an ML model with the publicly available data on $X_{ij}$ and $y_{ij}$, i.e., we aim to solve
\begin{align}
\label{eq:ml_train}
    \min \sum_{\forall (i,j)} \ell \big(\mathcal{F}(X_{ij}),y_{ij} \big).
\end{align}
%In Eq. \ref{eq:ml_train}, 
where $\mathcal{F}()$ is the output function of the ML model that takes values of $X_{ij}$ as input and the output is a representation of $y_{ij}$; the objective of the ML model is to minimize the difference between $\mathcal{F}()$ and the $y_{ij}$ values. The loss function $\ell()$ is one of the generalized loss functions, e.g., root mean square error. We expect that with enough training, the trained ML model will be able to predict the peering relationship on AS pairs not used in the training phase of the ML model. Fig. \ref{fig:framework} provides an overview of the proposed methodology with each block capturing the major components or tasks being done. In the following sections, we mainly focus on improving the implementation of the `Processing AS features' and `Training ML model' blocks shown in the figure to attain the minimum value with Eq. \ref{eq:ml_train} (`Training ML model' block).

{
\begin{remarks}
{\em The decision of whether two ASes should peer or not is inherently a binary decision, and hence our ML experimentation is a binary classification problem.}
\end{remarks}
}

\subsection{Datasets} \label{subsec:dataset} 
% \iia{Anindo, since you did not add info on the other databases, for now, I have skipped talking much about those and wrote the paragraph by just saying 1-2 lines on them. If possible may be try to add some more.} 

There are multiple publicly available datasets that offer a wide range of AS-level information in the form of infrastructure information, peering policies, routing policies, and AS relationships. This includes: PeeringDB, CAIDA, Packet Clearing House (PCH), Euro-IX, and Regional Internet Registry (RIR) databases (e.g., RIPE, APNIC) \cite{peeringDB_api, caida-as-rel, pch}. Among them, PeeringDB and CAIDA offer the most comprehensive insight into ASes. PeeringDB is maintained by the network operators themselves and provides granular information regarding traffic levels, peering policies, and geographical points of presence (PoPs), which are essential for accurately modeling peering relationships. CAIDA utilizes traceroutes, BGP routing tables, and historical AS paths to infer AS-level topology data, which provides significant insights into AS relationships (e.g., customer-provider, peer-peer). This can be utilized to determine the feasibility of potential peering relationships, making the combination of PeeringDB and CAIDA features an effective source for a peering partner prediction framework. On the other hand, datasets like PCH and Euro-IX primarily focus on IXPs and provide information about their connected networks. They lack AS-specific granularity and knowledge about peering policies to effectively determine peers. This is particularly problematic for Euro-IX which mainly focuses on mainland Europe and may result in erroneous predictions when used in other regions (due to the differences in the non-profit approach of European IXPs as opposed to the for-profit tendencies of US-based IXPs). RIR databases like RIPE and APNIC provide insight into AS-level registration characteristics and typically do not include operational metrics such as traffic patterns and peering policies.

% There are a few datasets that store information about ASes and are available to the public. PeeringDB and CAIDA are the most widely known for their inclusiveness of ASes and authenticity. We looked into a few more datasets, e.g., Packet Clearing house, EuroIX, ipdata, IPinfo, iamroot \cite{pch, euroix, ipdata, ipinfo, iamroot}; however, these datasets were not comprehensive in terms of ASes and corresponding features were also minimal. Hence, we decided to focus only on the PeeringDB (net) and CAIDA AS Rank data to extract the features of ASes \cite{peeringDB_api, CAIDA_AS_rank_api}. Alongside, CAIDA AS relationship data \cite{caida-as-rel} provides us with the peering relationship status between different AS pairs.

\begin{figure}[t]
    \centering
    \includegraphics[width=\linewidth]{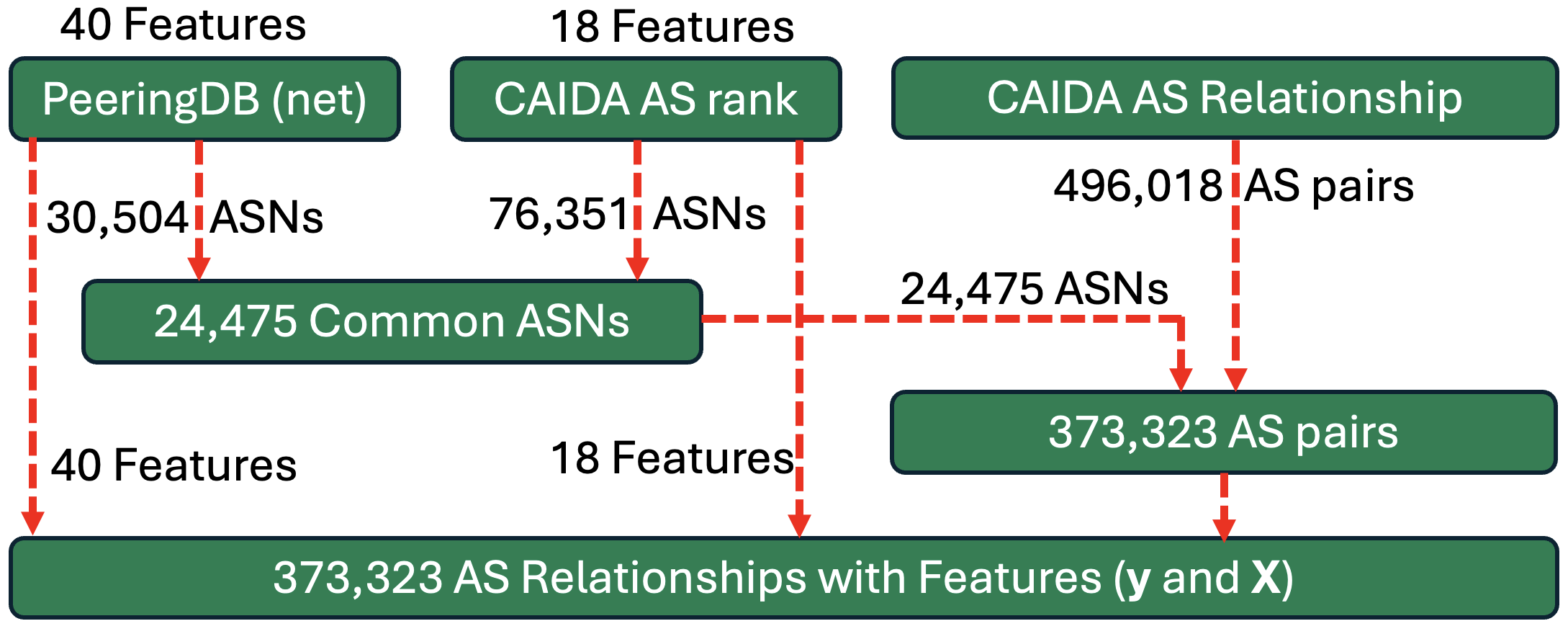}
    \caption{Data extraction and forming feature set of AS pairs.}
    \label{fig:data_extract}
\end{figure}

To retrieve the CAIDA AS rank data, we access CAIDA AS rank data v2 through API and download the total rank data of the ASes \cite{CAIDA_AS_rank_api}. We have downloaded the data from CAIDA in June 2024 and found that there are ranking data of 76,351 ASes. Some common features for each AS apart from their Autonomous System Number (ASN) were their rank, customer, peer, and provider numbers. The PeeringDB (net) data was downloaded from CAIDA as well\footnote{Although the data is available in PeeringDB, CAIDA has it in a more organized way.} \cite{CAIDA_public_data}. In the PeeringDB data from June 01, 2024, we find information on 30,504 ASes. The attributes of ASes available in PeeringDB data are very different compared to CAIDA AS rank data and include mainly AS traffic and policy-related information. Although CAIDA has almost 2.5 times more ASes than PeeringDB, we find that the 24,475 ASes common to both platforms are the most significant (in terms of ranking and customer cone) in the Internet ecosystem. Lastly, the CAIDA AS relationship dataset \cite{caida-as-rel} provides just under $500,000$ distinct peering relations between ISP pairs. In this dataset, the first two columns represent the ASNs, and the third column indicates whether a peering relationship exists between the two ASes (denoted by 0 for peering and -1 for non-peering). Fig. \ref{fig:data_extract} provides an overview of the datasets and how they were utilized.

% \begin{table}[t]
%     \centering
%     \begin{tabular}{p{2.4cm}|p{2.4cm}|p{2.2cm}}
%         Dataset Name & Info available (rows) & Features (columns)  \\ \hline
%         PeeringDB (net) & 30,504 ASes & 40 \\ 
%         CAIDA AS rank & 76,351 ASes & 18 \\ 
%         CAIDA AS relations & 496,018 AS pairs & 3 \\
%     \end{tabular}
%     \vspace{5pt}
%     \caption{Datasets}
%     \label{tab:data}
% \end{table}

\subsection{Data Processing}
\label{subsec:data_process}

\subsubsection{Feature Extraction}
\label{subsubsec:feature_extr}
There are in total 40 features per AS in the PeeringDB (net) dataset and among those 40, some features were noticeably insignificant to the decision-making of peering between AS pairs. Some of those features include the name of the AS, when the AS updated its information, the web link of the policy, etc. \newtext{Initially, we evaluate the performance using all the available features, and later we methodically omit features that are not useful for peering partner prediction.}
% We have removed those features from our feature set for each AS. For our initial dataset, we kept 18 features from the 40 (please refer to Table \ref{tab:data_used}). 
% On the other hand, 
\newtext{In the CAIDA AS Rank data, there are 18 features per AS, and similar to the PeeringDB data, some features seem irrelevant to the peering partner problem. However, for the initial evaluation, we perform peering partner prediction with all 18 features.}
% extracted 9 features to be included in the feature set of the ASes. 
\newtext{Hence, for each AS, we select $40$ features from PeeringDB and $18$ features from CAIDA, which gives us $58$ features\footnote{The full list of these features are given in Appendix \ref{appndx:features}.} in total. Among these features, $ASN$ is common for both, so we can count it only once. Additionally, in PeeringDB, each ASN is associated with a unique ID depending on the time it was registered. This ID can be considered as an alternate form of an ASN and can be excluded from the feature list of ISPs. As a result, the feature set consists of $58-2= 56$ features per ASN. Let us denote the set of these 56 features for some ASN $i$ as $f_i$.}

% have $40+18-1 = \mathbf{26}$ features (ASN is common to both datasets), and for AS $i$ we denote the set of these 26 features as $f_i$. Table \ref{tab:data_used} has the information of the features that are extracted from the full datasets.
% (also please refer to Fig. \ref{fig:data_extract} that has the whole process of the dataset generation)

% \begin{table}[t]
% \scriptsize
%     \caption{Features of the extracted datasets.}
%     \centering
%     \begin{tabular}{p{5.2cm}|p{2.65cm}}
%         \textbf{PeeringDB (net):}  &  \textbf{CAIDA AS rank:}\\ 18 features &  9 features\\ \hline
%         1) asn, 2) info\_ratio, 3) policy\_ratio, 4) info\_unicast, 5) policy\_general, 6) allow\_ixp\_update, 7) info\_types, 8) info\_traffic, 9) info\_multicast,	10) policy\_locations, 11) info\_scope, 12) fac\_count, 13) ix\_count, 14) org\_id, 15) policy\_contracts, 16) info\_prefixes6, 17) info\_prefixes4, 18) info\_ipv6.  & 1) asn, 2) total, 3) customer, 4) peer, 5) provider, 6) NumberASNs, 7) NumberPrefix, 8) NumberAddrs, 9) Rank.  \\
%     \end{tabular}
%     \vspace{5pt}
%     \label{tab:data_used}
% \end{table}

\subsubsection{Forming the Dataset for the ML Model}
%this paragraph is a bit confusing. there are disjointed bits of information here: the ASN relations dataset was discussed in section III.B but then it is explained in depth in Section III.C, which doesn't make sense.
Now that we have generated the feature set for the ASes, the next step consists of leveraging the AS relations dataset (if they are peering or not) as the blueprint to create a dataset that will be used as input to the ML model.
For each AS pair $(i,j)$, $X_{ij}$ is created by merging their respective feature sets, i.e., $f_i$ and $f_j$, and the value of $y_{ij}$ is collected from the AS peering relationship dataset of CAIDA (as discussed in Section \ref{subsec:problem_form}). There is information on $373,323$ AS pair peering relationships in CAIDA AS relationship dataset for the $24,475$ ASes that are common to both CAIDA AS rank and PeeringDB (net) AS feature sets. Hence, we have in total $373,323$ $X_{ij}$ and corresponding $y_{ij}$ values with each $X_{ij}$ having $\mathbf{56 \times 2 = 112}$ features. Please refer to Fig. \ref{fig:data_extract}, which delineates the process of data extraction and processing. 
Let us denote the full data of all the $X_{ij}$ values as $\mathbf{X}$, and all the $y_{ij}$ values as $\mathbf{y}$.
After compiling the data, all boolean features were converted to numerical values (True - 1, False - 0), and type mapping was applied to convert any other categorical features to numerical values (using Label Encoder)\footnote{\newtext{For performance measure with LLM (performed later), Label encoder was not used and the original textual values were used.}}.

\begin{remarks}
{\em There are $373,323$ AS pairs for which their peering status and pairwise features are publicly available.}
\end{remarks}

\subsection{Data Analysis}
\label{subsec:data_anlss}

To test the viability of using AS features to train an ML model and predict the AS relationship on unseen data, we first implemented a basic random forest (RF) model composed of 100 decision trees. We utilized the results of this model to evaluate the performance of the model as well as assess the importance of various features in predicting peering relationships. The results of this preliminary study are presented in this section.

%Revised paragraph:
%To test our hypothesis of using AS features to train a machine learning (ML) model for predicting AS relationships on unseen data, we first implemented a basic Random Forest (RF) Model. We utilized the results of this model to evaluate the performance of the model as well as assess the importance of various features in predicting peering relationships. The results for this preliminary study is discussed in this section.

% \begin{table}[t]
% \small
%     \centering
%     \begin{tabular}{p{1.88cm}|p{1.08cm}|p{1.08cm}|p{1.45cm}|p{1.08cm}}
%     Datasets used	& Overall Accuracy	& Balanced Accuracy & Non-Peering Accuracy	& Peering Accuracy \\ \hline
% PeeringDB 	& 95.88 &	89.95 &	81.13 &	98.76 \\
% % PeeringDB w/o ASN	& 95.53 &	89.12	& 79.59 &	98.65\\
% CAIDA AS rank	& 97.43 &	94.6 &	90.39 &	98.8\\
% % Caida  w/o ASN	& 97.32 &	94.45	& 90.18 &	98.72\\
% Both 	& 97.5 &	94.54 &	90.14 &	98.94\\
% % Both w/o ASN&	97.46 &	94.44 &	89.96 &	98.92
%     \end{tabular}
%         \vspace{5pt}
%     \caption{Accuracy (in \%) of RF based model for different training data.}
%     \label{tab:RAW_accr}
% \end{table}

\subsubsection{Performance Analysis}
\begin{figure}[t]
    \centering
    \includegraphics[width=0.95\linewidth]{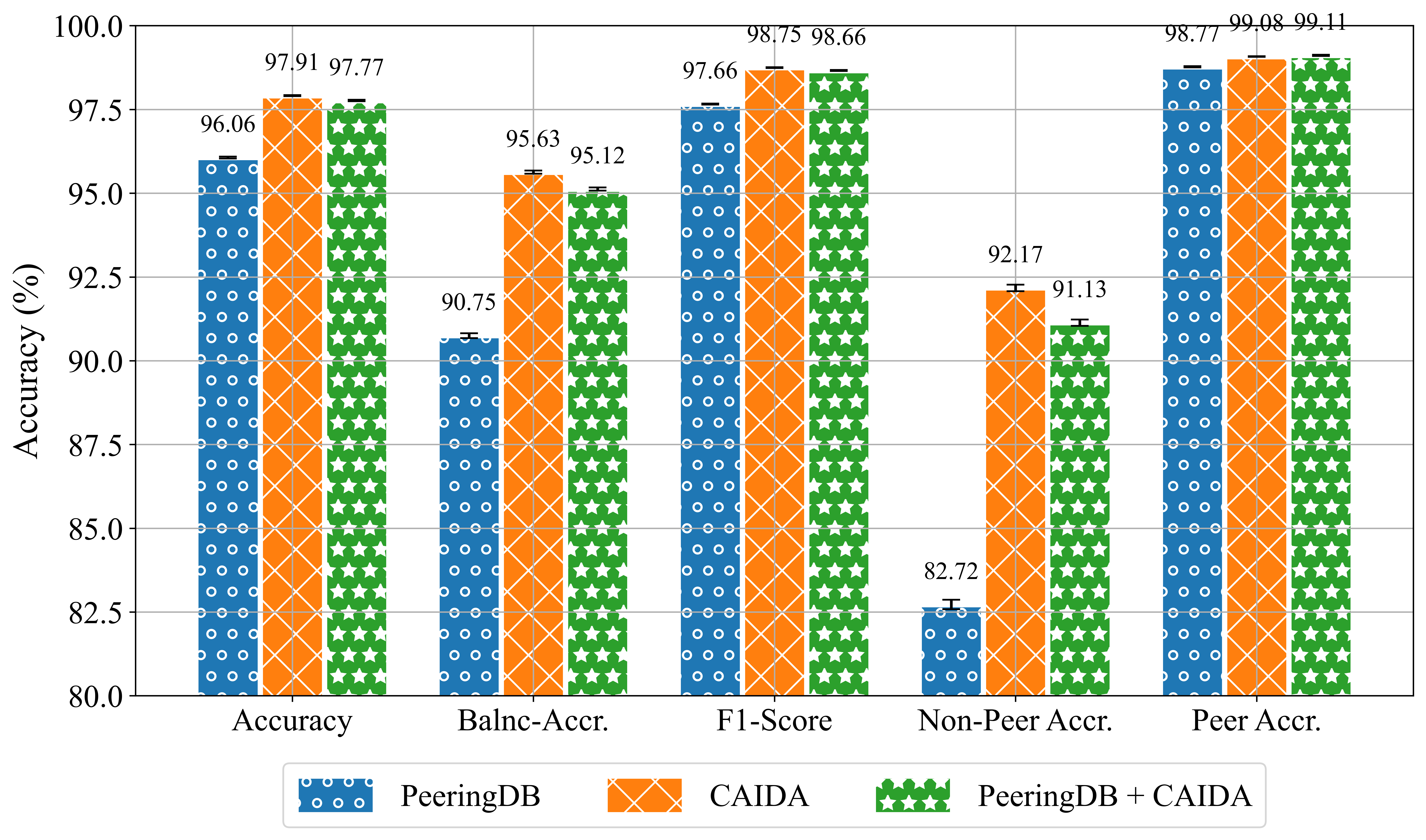}
    \caption{Accuracy (in \%) of RF based model for different training data.}
    \label{fig:raw_data_comp}
\end{figure}

For our initial evaluation, the processed data ($\mathbf{X}$ and $\mathbf{y}$, see Section III - \ref{subsec:data_process}) were randomly split in a 70:30 ratio using the train-test-split function \cite{sklearn_train_test_split}. That is, $70\%$ of the rows (from $\mathbf{X}$ and $\mathbf{y}$ with matching $X_{ij}$ and $y_{ij}$) were randomly chosen and placed into the training set, while the other $30\%$ is placed into the test set. The training data is then sent as input to the RF classifier, while the test data is used to benchmark the accuracy of the model.
% The goal of supervised learning is to build a model that is robust with new data. Train-test-split is an effective model validation process that properly simulates ML model predictions with unseen data. 
% The process focuses on splitting a set of data into two smaller sets - training and testing. 
% For this experiment, we chose a ratio of 70:30, which means that }
%Table \ref{tab:RAW_accr} 
Fig. \ref{fig:raw_data_comp} shows the accuracy values attained using the RF model for different datasets with a train-test-split of 70:30. Since we have utilized 2 datasets (PeeringDB net and CAIDA AS Rank) for AS features, we applied all possible combinations of features as the input to our RF model. For each scenario, the recorded metrics included the F1-score, as well as the overall, balanced, non-peering, and peering accuracy\footnote{The peering accuracy is the accuracy of the model in identifying AS pairs to be in peering relationship when they are actually peering (according to CAIDA AS rank). Non-peering accuracy is defined in the same way, but it is when the AS pairs are not peering and how accurately the model can predict that. Lastly, balanced accuracy is the average of non-peering and peering accuracy.}. \newtext{This evaluation and all subsequent simulations (unless explicitly mentioned) are conducted using 20 different random seeds, where the average value is recorded. The black vertical lines in the figure represent one standard deviation from the average value.} 
% with a confusion matrix, and balanced accuracy by taking the mean of non-peering and peering accuracy. 
% Each variation of the dataset (PeeringDB, CAIDA, and PeeringDB+CAIDA) was inputted into our RF model twice. Since both datasets have ASN as a feature, the first trial was with ASN as a feature in the dataset while the second trial was without ASN as a feature. 
From the figure, we observe that although PeeringDB has a relatively high overall accuracy, there is a large gap between the non-peering and peering accuracy (leading to a lower balanced accuracy), which is not ideal. 
% Apart from the ASN feature, the PeeringDB and CAIDA train dataset has 17 features and 8 features {\em per ASN} respectively.
Interestingly, even with 
\newtext{almost half the number of features,}
% a minimal amount of 8 features, 
the CAIDA AS Rank dataset performed exceptionally (orange bars) compared to PeeringDB (blue bars). Lastly, merging both data sets \newtext{does not yield significant benefits. Instead, we observe a $1\%$ decrease in the balanced accuracy. This result indicates that there might be many features in PeeringDB that may negatively impact performance and also emphasizes the importance of the CAIDA AS Rank data.} 

\subsubsection{Feature Importance and Shapley Values}
\label{subsec:feature_and_shap}

\newtext{To evaluate the importance of each feature in the peering partner prediction process, we first calculate the feature importance scores using the mean decrease impurity method. For completeness, we include all 56 features of the ISP, including $ASN$ as well. The feature importance results are provided in Appendix \ref{appndx:feature_imprtnc}. To further support the evaluation, we also perform a Shapley value analysis. The details of the results are also provided in Appendix \ref{appndx:feature_imprtnc}. Based on these analyses, we observe that 15 features have negligible or slightly negative effects on peering partner prediction. Hence, we remove these features from the AS feature set for all ASes, resulting in the feature set $f_i$ for AS $i$ to consist of $41$ $(56-15)$ features. Moreover, we observe that a few features (around 20) are the most important ones in deciding ISP peering relationship and some of these features, such as ASN, play an unexpectedly important role. Future work on the theoretical developments in this topic can use this insight to build models.}

% , which is in agreement with what we saw in the results shown in Fig. \ref{fig:FI_drop_accr}. 

% \begin{figure}[t]
%     \centering
%     \includegraphics[width=0.9\linewidth]{figures/Feature_importance_MDI.png}
%     \caption{Feature importance (mean decrease impurity).}
%     \label{fig:FI_1}
% \end{figure}

\subsubsection{Correlation between Features}

\begin{figure}[t]
    \centering
    \includegraphics[width=0.95\linewidth]{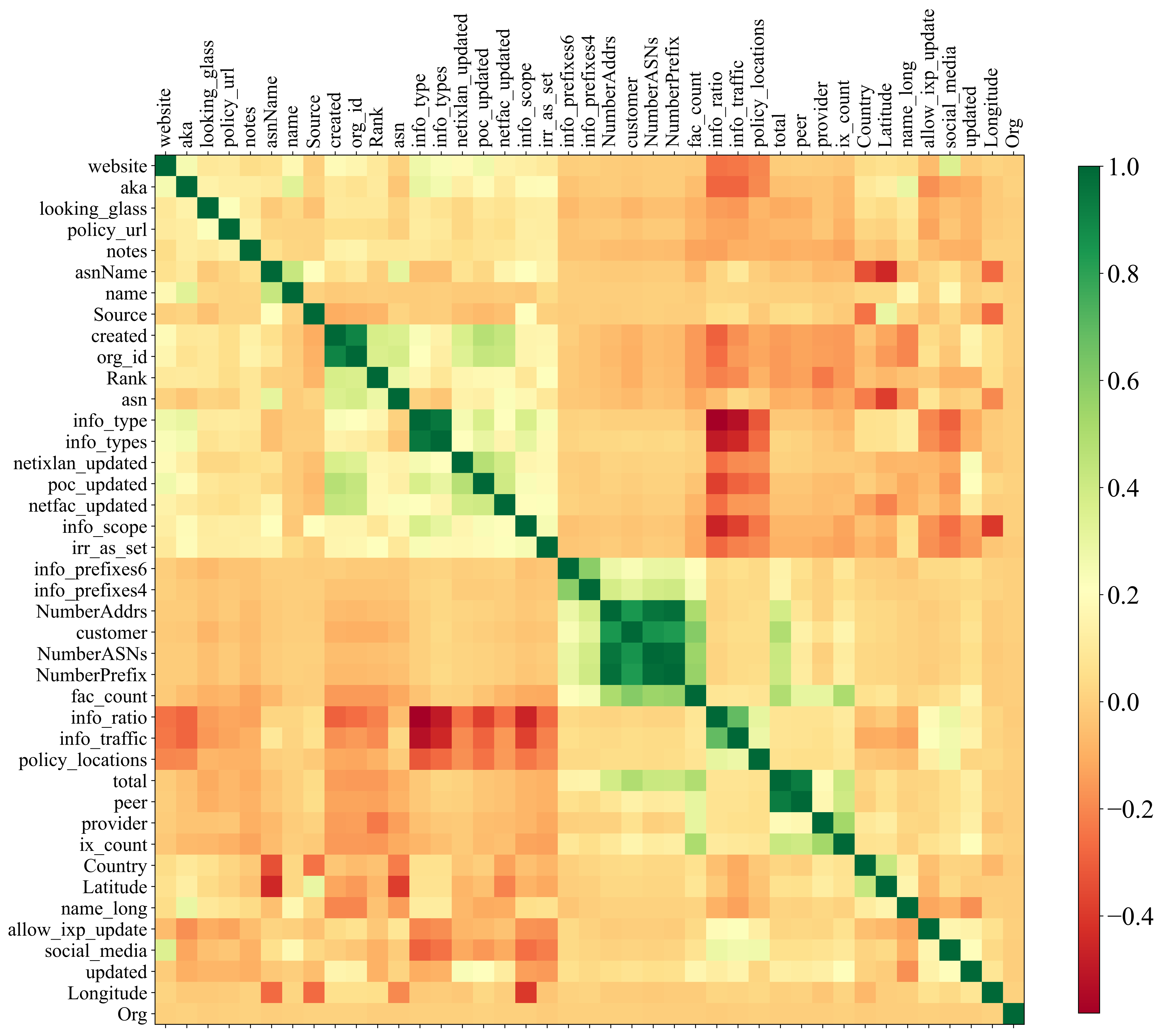}
    \caption{Correlation between different features.}
    \label{fig:correlation}
\end{figure}

The correlation between the \newtext{41} features (per AS) is shown in Fig. \ref{fig:correlation} \cite{scipy_hier_cluster}. In general when training an ML model, if multiple features are highly dependent (correlated) to one another, then instead of utilizing all those dependent features in the feature set, only one (or few) feature(s) can be used without sacrificing much performance. From the correlation matrix shown in Fig. \ref{fig:correlation}, we see that there are a few features that are highly correlated with one another. 
\newtext{We observe some high correlations between i) $NumberASNs$, $NumberPrefix$ and $NumberAddrs$ features, ii) $Peer$ and $Total$ features, iii) $info\_type$ and $info\_types$ features, and iv) $created$ and $org\_id$ features. These observations suggest that it might be possible to decrease the feature set by using one or a few correlated features instead of all $41$ feature pairs.}

\subsubsection{Effect of Features on Performance}

\begin{figure}[t]
    \centering
    \includegraphics[width=\linewidth]{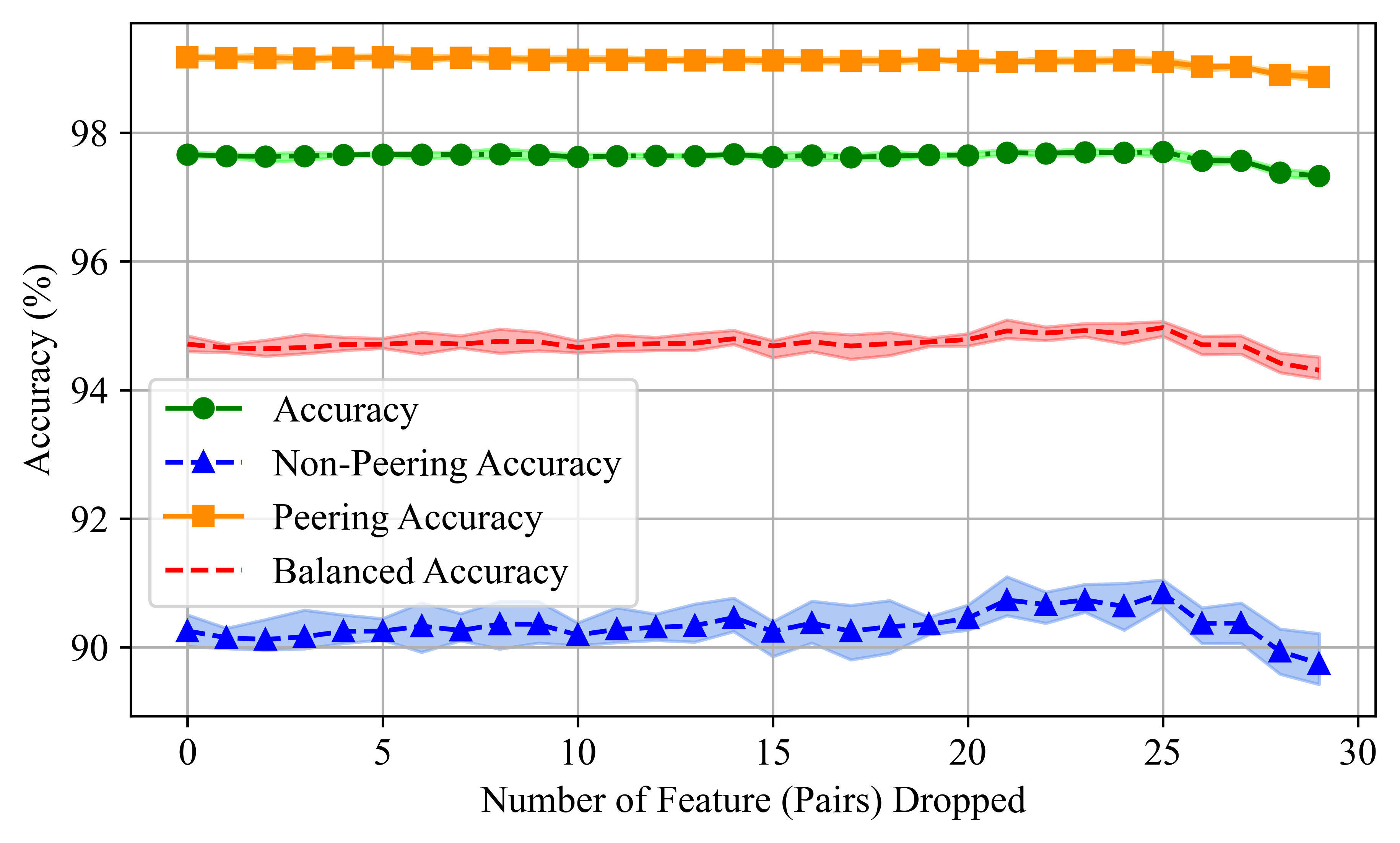}
    \caption{Accuracy of RF model by sequentially dropping least important feature (pairs) - 30 pairs.}
    \label{fig:FI_drop_accr_least30}
\end{figure}

\newtext{To further investigate the impact of the features on model performance, we sequentially remove the least important features from the feature set $\mathbf{X}$ (currently $41+41=82$ features), and retrain the RF model after each removal to evaluate performance. The results of this experiment are provided in Figs. \ref{fig:FI_drop_accr_least30} and \ref{fig:FI_drop_accr}. The entire process was repeated 20 times with different seeds to ensure consistency. The list of features in terms of their least importance (as removed one by one) is provided in Appendix \ref{appndx:drop_featrs_1by1}. 
%were: 1) peer, 2) customer, 3) NumberAddrs, 4) Rank, 5) total, 6) NumberASNs, 7) NumberPrefix, 8) info\_prefixes4, 9) ix\_count, 10) info\_prefixes6, 11) fac\_count, 12) provider, 13) asn, 14) org\_id, 15) info\_traffic, 16) policy\_general, 17) info\_scope, 18) policy\_locations, 19) info\_ratio, 20) policy\_contracts, 21) info\_types, 22) policy\_ratio, 23) allow\_ixp\_update, 24) info\_multicast, 25) info\_ipv6, 26) info\_unicast. 
Fig. \ref{fig:FI_drop_accr_least30} shows the performance when we drop 30 of the least important features (the full result is in Appendix). 
The results indicate that the 20 least important features contribute little to improving peering prediction. Furthermore, there is a slight performance gain observed when removing up to the 25th least important feature. Therefore, the 16 ($41-25$) most important features appear to form the optimal feature set for the RF model, achieving high performance with a minimal number of features\footnote{ detailed performance comparison between the top 16 and top 41 features is provided later in the Ablation Study (Section \ref{sec:ablation-study})}.}

\begin{remarks}
{\em Only a few features have significant influences on the peering relationship of two ASNs.}
\end{remarks}

% When removing from most important to least: 1) peer, 2) total, 3) customer, 4) NumberAddrs, 5) NumberASNs, 6) Rank, 7) NumberPrefix, 8) info\_prefixes4, 9) info\_prefixes6, 10) provider, 11) asn, 12) ix\_count, 13) fac\_count, 14) org\_id, 15) info\_traffic, 16) policy\_general, 17) info\_scope, 18) info\_types, 19) policy\_locations, 20) policy\_contracts,  21) policy\_ratio, 22) info\_ratio, 23) allow\_ixp\_update, 24) info\_ipv6, 25) info\_multicast, 26) info\_unicast

% \begin{figure}
%     \centering
%     \includegraphics[width=\linewidth]{figures/Accuracies_by_removing_Important_feature_pairs.png}
%     \caption{Accuracy of RF model by sequentially dropping most important feature (pairs).}
%     \label{fig:FI_drop_important_accr}
% \end{figure}

% \input{tex_files/4_Data_analysis_and_model_selection}
\section{Model Selection and Ablation Study}
\label{sec:ablation-study}
From Section \ref{sec:data}, we have observed that some of the features in the extracted data are more important in the decision-making of peering partners. Hence, it is imperative to perform an ablation study and check if we can attain an improved and more concise set of features that performs close to the original dataset with all features. However, before moving to that, let us focus on selecting the optimal ML model for peering partner prediction. 

\subsection{Model Selection}
There are a few well-known and popular ML models available for supervised learning. Since we do not know which will be a better suit for our case, we explore a few major types, i.e., support vector machine (SVM), tree-based methods (RF and XGBoost), neural networks (e.g., MLP, DNN), and transformer-based models \cite{muhamedyev2015machine, vaswani2017attention}. For SVM, we used a hyperparameter of max\_iter = 10000; because the model was not converging. Since RF and XGBoost (or XGB in short) perform close to their optimal in default settings, we did not perform hyperparameter optimization. For the neural network model, we went with a 3-layer model that has 100 neurons in each layer and the output layer has one neuron for peering prediction\footnote{We have found that increasing the number of layers or neurons per layer did not improve the performance and the multilayer perceptron (MLP) also did not perform well.}. For the sake of simplicity, we will call this neural network model a deep neural network (DNN) onward. Lastly, for transformer-based models, we choose a \newtext{longformer (long transformer) model\footnote{The pre-trained model that we used is `allenai/longformer-base-4096'}. The reason behind using this model is that when we use all the features (including features with text), the full input text becomes too large (more than 3000 tokens) to be supported by BERT, BART, and similar models.
% BERT base model. Since BERT is quite computationally expensive compared to all the other methods explored here, we refrained from going to an even larger model. 
}

\begin{table}[t]
\caption{Performance comparison of different ML models.}
\small
    \centering
    \begin{tabular}{p{0.9cm}|p{1.08cm}|p{1.08cm}|p{0.8cm}|p{1.1cm}|p{1.1cm}}
    Model Name	& Overall Accuracy	& Balanced Accuracy & F1 Score & Training Time (sec) & Evaluation Time (sec) \\ \hline
DNN  & 96.181 &	93.831 & 97.695 & 24.949 &	 3.647\\
RF	& 97.375 & 94.196 & 98.43 & 24.917 & 2.63	\\
SVM	& 95.56 & 90.097 & 97.348 &  210.5 & 503.1 \\
XFMR & 96.222 &	92.985 & 96.211 & 18740 & 19200 \\
\textbf{XGB}	& \textbf{97.593} &	\textbf{94.935} & \textbf{98.558}	& \textbf{22.57} & \textbf{0.132 }
    \end{tabular}
        \vspace{5pt}
    \label{tab:model_compr}
\end{table}

Table \ref{tab:model_compr} compares different performance attributes of the five models trained to perform peering partner prediction, \newtext{where the transformer-based model is denoted as XFMR. For this evaluation specifically, all models are trained using the full set of 56 features per AS to ensure consistency and to allow the transformer model to effectively utilize the textual data.} 
Since the transformer model is computationally heavy compared to all the other models, and we have quite a large training data, we used $25\%$\footnote{\newtext{{\em Six V100 GPUs} were used to speed up the training and evaluation time of the transformer model, but we could still only train using 25\% of the data.}} of the whole data for training and the remaining $75\%$ for performance testing (for all models). \newtext{Upon close observation, we see that the transformer based model (XFMR) did not perform well. Moreover, due to it being so computationally expensive, the training and evaluation times were also quite long.}
Overall, in terms of performance, SVM yielded the weakest results, while DNN performed closely to RF and XGB. In general, tree-based methods prevailed in performance and speed, with {\em XGB} being the best model in our experimental setting. \newtext{This result is not too surprising due to several reasons: a) tree-based models generally perform better on tabular data \cite{shwartz2022tabular, zabergja2024deep}; b) when data is abundant, tree-based models often outperform neural network-based models \cite{mcelfresh2023neural}, i.e., DNN and transformers; and c) tree-based models are typically more robust to asymmetric data or skewed feature distributions \cite{uddin2024dataset, mcelfresh2023neural}.}

\begin{remarks}
{\em
For the rest of the paper, unless stated otherwise, we will use the XGBoost model for performance analysis.}
\end{remarks}

\subsection{Data Sampling}
\begin{figure}[t]
    \centering
    \includegraphics[width=\linewidth]{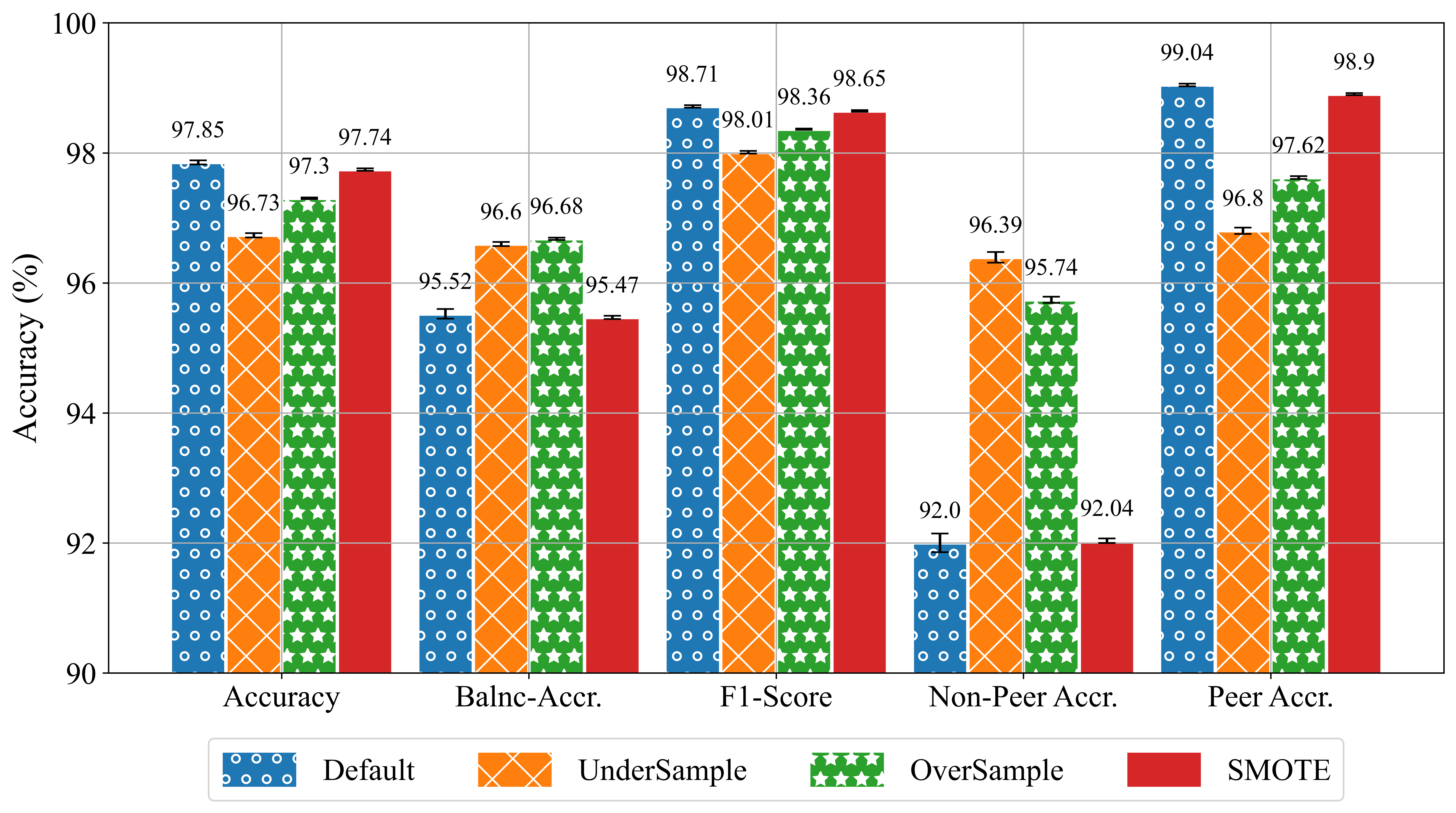}
    \caption{Performance comparison with over and undersampling.}
    \label{fig:over_under_comp}
\end{figure}

The CAIDA AS relationship data is heavily imbalanced in the peering status of AS pairs. Among the $373,323$ AS pair data that we extracted, $309,396$ AS pairs are labeled as peering partners, while the rest $63,927$ are labeled as non-peering partners (a ratio of $83:17$). Oversampling and undersampling are two common methods used in ML when datasets are imbalanced. In oversampling, the minority class data is copied until it becomes equal to the size of the majority class data. In undersampling, we remove data from the majority class randomly until the data size decreases to the size of the minority class. The results of experimenting with oversampling and undersampling are depicted in Fig. \ref{fig:over_under_comp}\footnote{Oversampling and undersampling is done only to the training dataset, not on the test data.}. \newtext{The figure also includes results for SMOTE, a commonly used technique for addressing data imbalance \cite{pradipta2021smote}. It is evident that both data undersampling and oversampling can increase non-peering accuracy, ultimately increasing the balanced accuracy. However, the overall accuracy and F1-score are higher with the original dataset and SMOTE. Our understanding of the Internet ecosystem suggests that there is no universally optimal evaluation metric—such as accuracy, balanced accuracy, or F1-score—as ideally, all should be maximized. The decision of which metric to prioritize and the application of sampling techniques to optimize that metric may depend on the specific characteristics and goals of each ISP pair, making it quite subjective.}
% Although we expected either of the two methods (over- or undersampling) to perform better than using the original (training) dataset, that was not the case. Peering accuracy with under- and oversampling is higher compared to the default data; however, the non-peering accuracy is poor, resulting in less accuracy and balanced accuracy. 
Upon this observation, we have decided to use our default dataset without performing any over or undersampling for the rest of the paper\footnote{\newtext{It might be feasible to train two different models: one with the default dataset and the other with data oversampling to serve different service criterion of the ISP pairs.}}.

\subsection{Ablation Study}

The current dataset we have extracted from the CAIDA AS Rank and PeeringDB net, has $82$ \newtext{($41+41$, after excluding the 15 features per AS, as described in Section \ref{sec:data} - \ref{subsec:data_anlss}.\ref{subsec:feature_and_shap})} features per AS pair and the corresponding label (output) stating if they are peering (extracted from CAIDA AS relationship). Let us call this extracted dataset as the `\emph{default dataset}' (with slight misuse of the term\footnote{This `default' dataset is a filtered and processed version of a larger feature-set, as discussed in Section \ref{sec:data} - \ref{subsec:data_anlss}.\ref{subsec:feature_and_shap}.}). From the discussion in Section \ref{sec:data} we know that there might be redundancy in the default dataset and that some of the features can be dropped with almost no performance drop. Moreover, the work in \cite{mustafa2022peer} inspired us to look into two new features, namely `Cone Overlap' and `Affinity Score', and the possibility of a better-processed dataset that may attain better performance.

From the discussions above, we have devised three distinct methods to generate or modify the default dataset and create three new datasets for performance comparison. Let us provide short descriptions of the three new datasets along with the default dataset before moving on with their performance analysis;

\begin{enumerate}
    \item \textbf{Default Dataset:} The default dataset has \newtext{$82$} features per AS pair (i.e., \newtext{$41$} features per AS) and is generated with the process described in Section \ref{sec:data} - \ref{subsec:data_anlss}.\ref{subsec:feature_and_shap}. 
    \item \textbf{Filtered Dataset:} The `filtered dataset' is a filtered version of the Default Dataset and the top $16$ features per AS is picked from the $41$ features of the default dataset. The $16$ features are: 1) $customer$, 2) $peer$, 3) $NumberAddrs$, 4) $NumberPrefix$, 5) $total$, 6) $Rank$, 7) $NumberASNs$, 8) $info prefixes4$, 9) $ix count$, 10) $info prefixes6$, 11) $fac count$, 12) $provider$, \newtext{13) $asn$, 14) $Latitude$, 15) $created$, and 16) $Longitude$. Thus, the filtered dataset has $32$} features per AS pair. 
    \item \textbf{Processed Dataset:} The `processed dataset' is generated by mainly following the procedure in \cite{mustafa2022peer}. In that work, they performed some simple algebraic processing of the features to reduce the total number of features and also introduced two new features (`$Cone Overlap$' and '$Affinity Score$') that are not available in the CAIDA AS Rank or PeeringDB net datasets. The full procedure of generating this dataset is provided in the following section. Overall there are $17$ features per AS pair, among which `$Cone Overlap$' and '$Affinity Score$' are generated using data outside of PeeringDB net and CAIDA AS rank \newtext{(this data is also available in CAIDA, but in a different format)}.
    \item \textbf{Optimum Dataset:} We are calling the `optimum dataset'\footnote{We will observe later that this dataset attains (near) optimum performance, i.e., better performance with small features set} as the one where we add `$Cone Overlap$' and '$Affinity Score$' (these two features are not for each AS, rather for each AS pair) features to the `Filtered Dataset' and coming up with a $32+2 = 34$ features per AS pair dataset.
\end{enumerate}

\begin{figure}[t]
\centering
\begin{minipage}{0.49\linewidth}
\centering
\includegraphics[width=\linewidth]{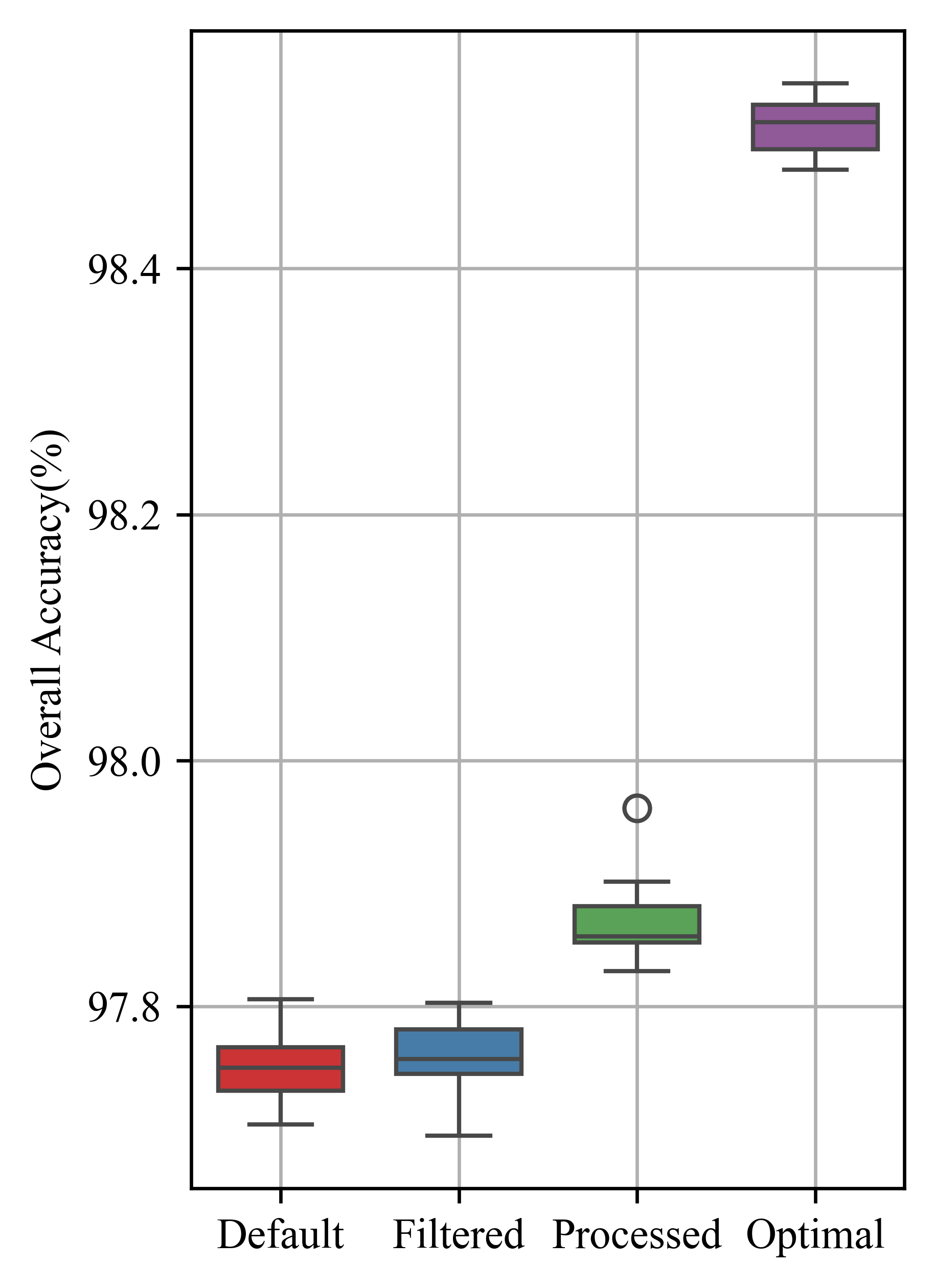}
\caption{Overall accuracy.}
\label{fig:4_methods_accr}
\end{minipage}
\hfill
\begin{minipage}{0.49\linewidth}
\centering
\includegraphics[width=\linewidth]{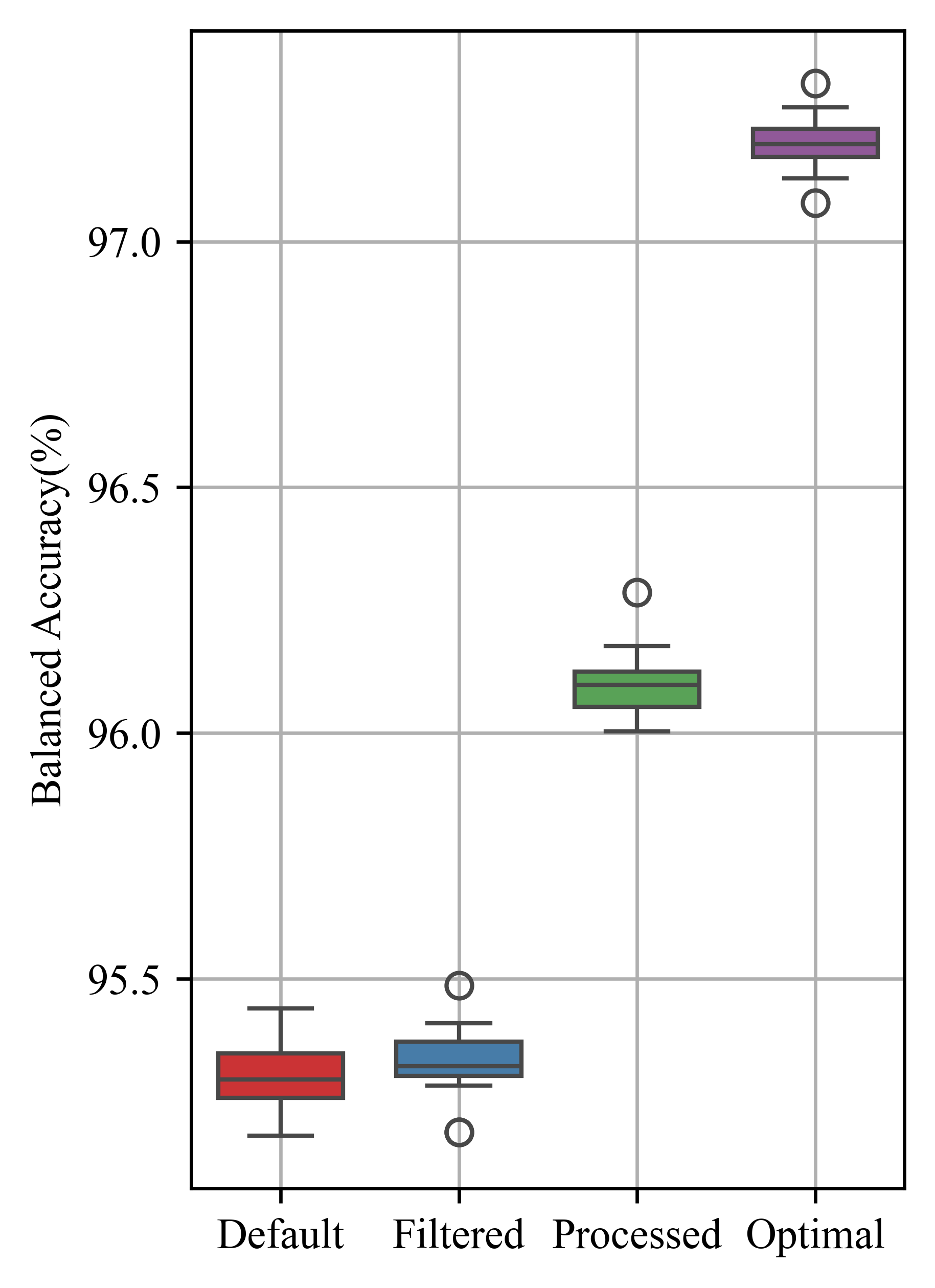}
\caption{Balanced accuracy.}
\label{fig:4_methods_b-accr}
\end{minipage}

\centering
\begin{minipage}{0.49\linewidth}
\centering
\includegraphics[width=0.98\linewidth]{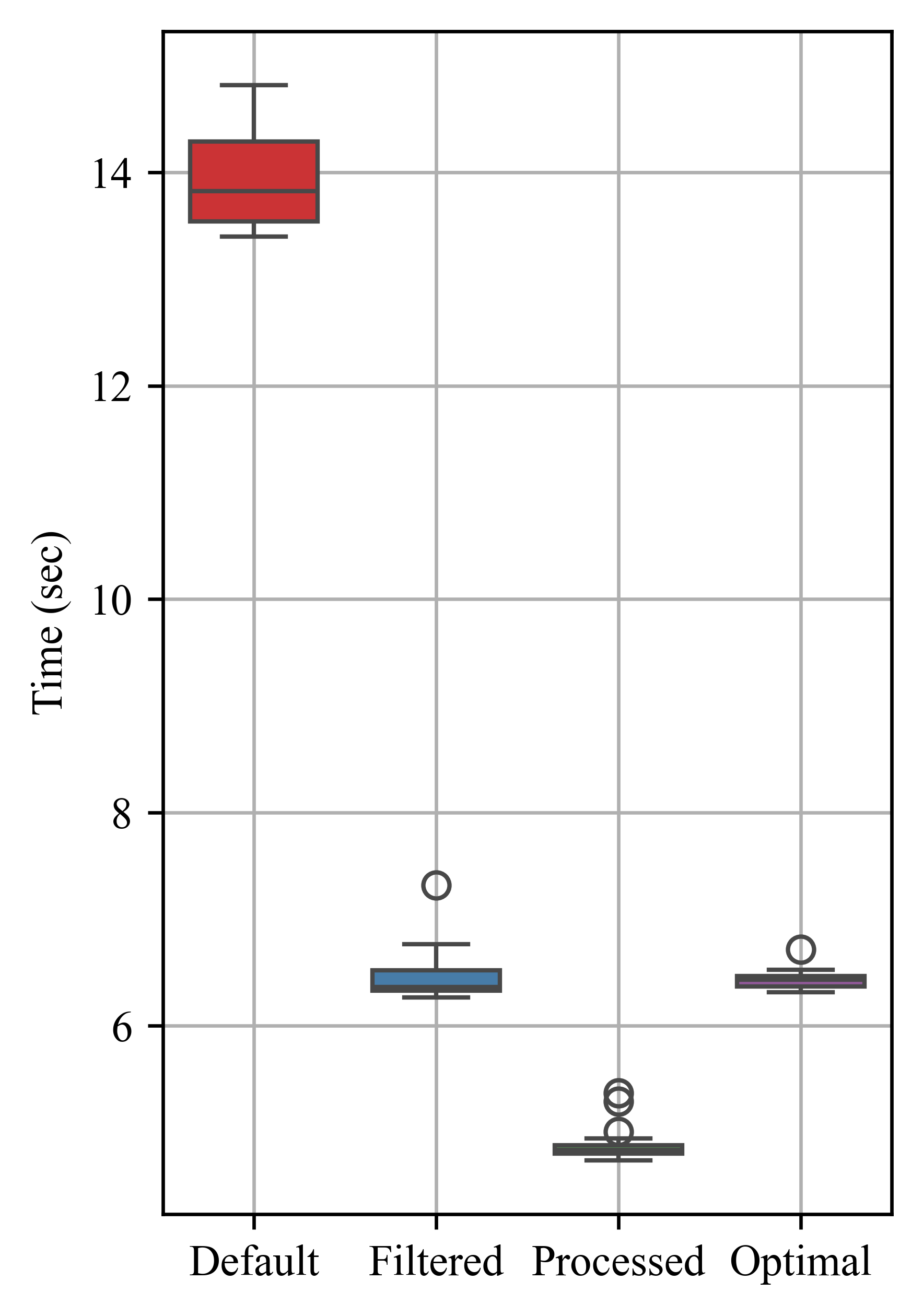}
\caption{Fit time.}
\label{fig:4_methods_fit_time}
\end{minipage}
\hfill
\begin{minipage}{0.49\linewidth}
\centering
\includegraphics[width=\linewidth]{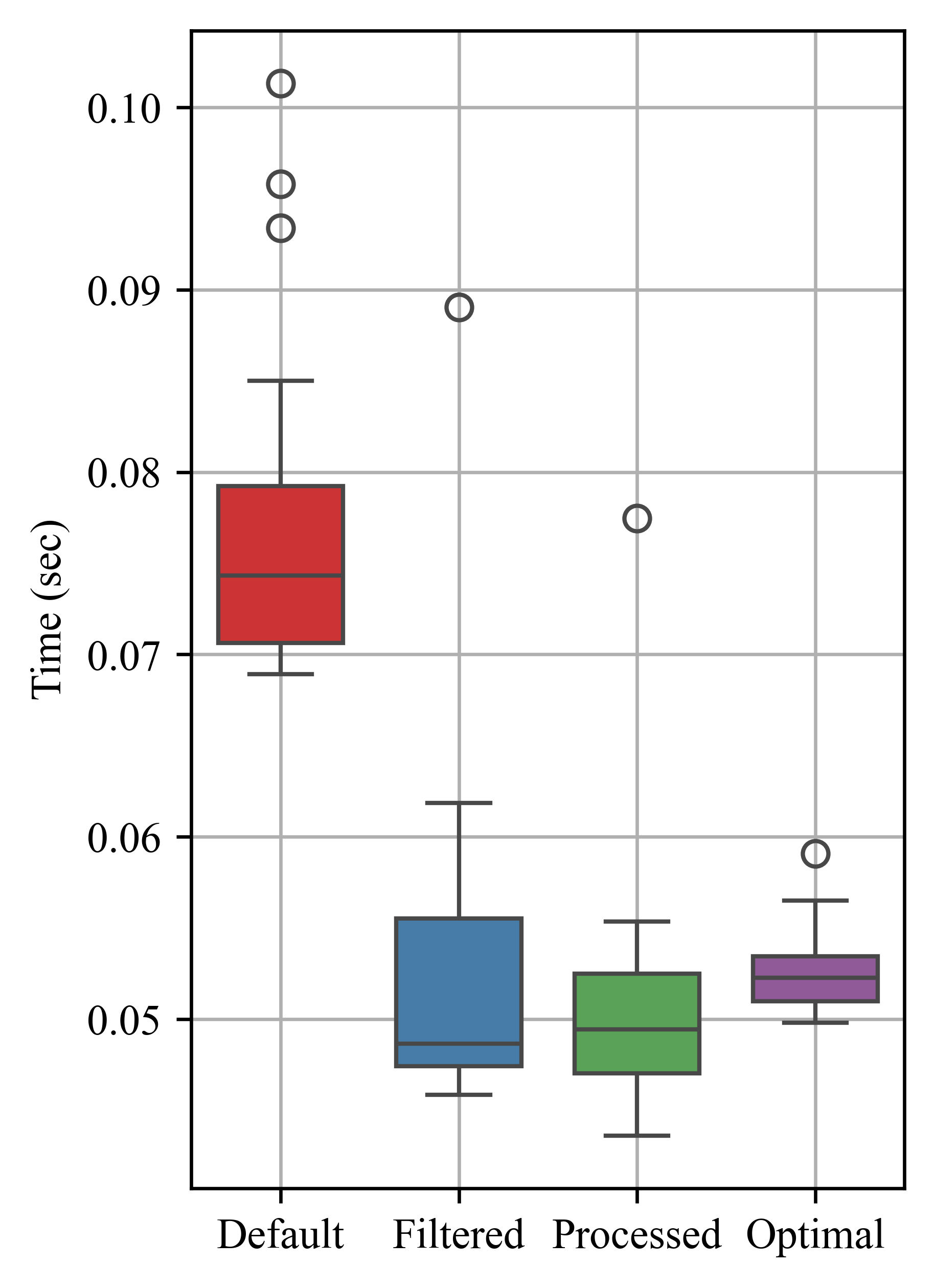}
\caption{Evaluation time.}
\label{fig:4_methods_eval_time}
\end{minipage}
\end{figure}

Figs. \ref{fig:4_methods_accr} and \ref{fig:4_methods_b-accr} respectively show the overall and balanced accuracy achieved when using the four datasets discussed above to train the ML model (XGBoost). For the model training, we used $50\%$ of the data to train the model and tested on the rest $50\%$ data. We did this 20 times with 20 different seeds to check the consistency of performance for all four datasets; the box plots show the inter-quartile ranges of the results. The model fit-time (training time) and evaluation time (on test data) for these four datasets are shown in Figs. \ref{fig:4_methods_fit_time} and \ref{fig:4_methods_eval_time}, respectively. We note that the Filtered Dataset, with only $32$ features, gives an accuracy (overall and balanced) better than the Default Dataset of $82$ features; thus saving memory by more than $60\%$ and training time by $50\%$. 

Interestingly, when we move to the Processed Dataset, the performance becomes even better than the Default and Filtered Datasets with only $17$ features per AS pair. Since the Processed Dataset mainly consists of data from PeeringDB (net) and CAIDA AS rank, apart from the two features calculated from other data sources, we speculate that these two features should have contributed to the performance improvement. Our speculation was proven right when we checked the performance of the Optimal Dataset, where these two new features (`$Cone Overlap$', and `$Affinity Score$') were added to the Filtered Dataset. From Fig. \ref{fig:4_methods_accr} to Fig. \ref{fig:4_methods_eval_time}, we observe that the Optimum Dataset has the best performance \newtext{(approximately 1\% accuracy improvement)} along with a short training and evaluation time, which is comparable to the times with the Processed Dataset. From these observations, we have decided to use only the Optimal Dataset to perform the robustness study of the ML model in the next section.

\begin{remarks}
{\em There is a feature-set that ensures close to maximum accuracy with small number of features. We denote this dataset as the Optimal Dataset.}
\end{remarks}

The Processed dataset (and Optimum dataset) has two features that are not directly available from PeeringDB and CAIDA datasets, namely; Cone Overlap and Affinity Score. The procedure of how those were calculated is as follows:

\noindent \emph{Cone Overlap and Affinity Score in Calculation:}

To calculate the cone overlap for an AS pair we follow the procedure used in \cite{mustafa2022peer}. We start by constructing the customer cones of each of the ASes using the CAIDA AS relationships dataset. The customer cone of an AS includes all the ASes that are customer to that AS or within the cone of the customer ASes.
% Usually peering among two ASes does not allow traffic transit
% between their indirect customers. An ASes indirect customers
% refer to the customers of its customers. This means that if
% AS A and AS C are peering, indirect customer ASes of
% A will not be able to reach indirect customer ASes of C
% using the peering link as a transit route. If a provider AS A
% starts peering with one of its customers AS C, its customer
% cone size will reduce (see Figure 8). Therefore, 
For an AS
pair, the cone overlap is the total number of common ASes that are present in the customer cones of both ASes. 
% We refer to this as the Cone
% Overlap and expect it to play a significant role in peering
% decisions of ASes.
Calculation of the affinity score
% An AS will be interested in peering if the
% relationship would expand its coverage area; otherwise, there
% may not be enough incentive to peer with an AS that is
% covering the same locations or has a smaller coverage area.
for a given AS pair is also done by following the method presented in \cite{mustafa2022peer}. We start by calculating the PoP affinity of each AS (say AS 1) to the other AS (say AS 2):
\begin{align}
    \alpha_{1\to2} = \frac{P_2-P_0}{P_1 \cup P_2} = \frac{P_2-P_0}{(P_1-P_0)+(P_2-P_0)+ P_0},
\end{align}
where $P_1$ and $P_2$ are the number of PoPs for ASes $1$ and
$2$ respectively, and $P_0$ is the number of their common PoPs.
Then the affinity score for the AS pair $(1,2)$ is calculated as:
\begin{align}
    \alpha_{1,2} = \sqrt{\alpha_{1\to2}\times \alpha_{2\to1}}.
\end{align}

% We use
% geometric mean to calculate the combined PoP Affinity

% The geometric mean assures that both ASes A and C will
% increase their coverage if they peer. A situation where only
% A or C increases its coverage from peering would not be
% desirable since only one AS would benefit from peering in
% that case. As expected, PoP affinity AC has a positive impact
% on the probability of peering

\subsection{Performance Comparison to Other Works} 
\begin{figure}[t]
    \centering
    \includegraphics[width=0.95\linewidth]{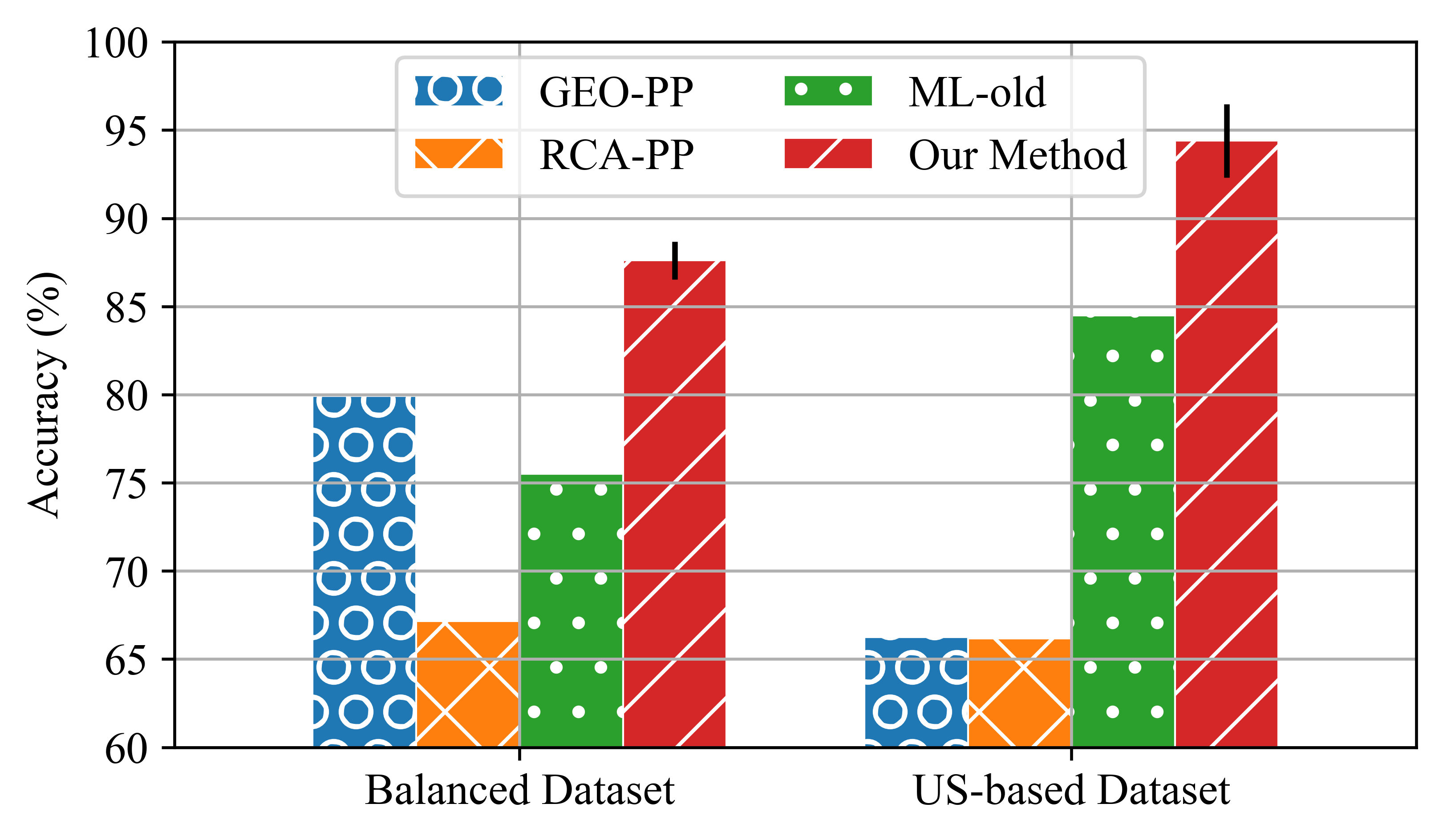}
    \caption{Performance comparison with other works.}
    \label{fig:comparison_others}
\end{figure}

Peering partner prediction on a global scale has been previously explored in \cite{alam2024meta, mustafa2022peer}. To compare the performance of our method with the methods proposed in those works, we \newtext{focused on two aspects, i) dataset consistency and ii) ML training method consistency. At first,} we created the two datasets consisting of the ASes used in \cite{alam2024meta}, namely a) balanced dataset, and b) US-based dataset. For these two datasets, we implemented our \newtext{methodology to generate the feature sets and train an {\em RF model} with 70\% of the data, and then test the performance on the remaining 30\% {\em (to be consistent with the ML training method)}}. We ran for 10 runs with different seeds to check performance consistency. The results are shown in Fig. \ref{fig:comparison_others} with the standard deviation of our results in a black vertical line (the values of the other methods are directly extracted from \cite{alam2024meta} and hence the standard deviation line is missing). From the figure, it is evident that our method is superior in terms of predicting peering relationships between AS pairs. The ML-old is the method proposed in \cite{mustafa2022peer}, whereas RCA-PP and GEO-PP were proposed in \cite{alam2024meta}, and our method outperforms the best methods by more than 6\% accuracy margin for both datasets. 

\subsection{Potential Partner Prediction: Unknown Peering Status}
One of the main objectives of this work is to use our proposed method to find potential peering partner ISPs who are not already peering with each other. To check such cases, we created a dataset using 20 larger ASes as one peering partner and checking their peering partnership status with other ASes for which no data is available in the CAIDA AS-relationship dataset. We looked into $469,921$ AS pairs and predicted their peering status (what it should be) with our model (fully trained on the currently available data from the CAIDA AS relationship). The model output suggested that only $33.51\%$ of the pairs are good for peering relationships and the rest are not. This is interesting because the CAIDA AS relationship dataset is heavily biased with 83\% of the AS pairs being in a peering relationship; whereas, for this unknown peering status dataset we are getting much fewer potential AS pairs to peer. It would be interesting to see if these AS peers actually pair up in the future. On that note, one such approach is explored in the next section, where we trained our model with data from 2022 and checked on data from 2024; the results are quite promising.

\section{Peering Partner Prediction - ML Performance Robustness}
\label{sec:robustness}

The results in the previous section bolster the proposition of an ML-based approach to predict the peering partner of ASes. Although the results are promising, we do not know how robust the ML models are, e.g., how much training data is needed, whether the model is versatile over time, etc. To evaluate the robustness of this ML-based technique, we perform a few robustness tests in this section.

\subsection{Training Data Size}

\begin{figure}[t]
    \centering
    \includegraphics[width=0.95\linewidth]{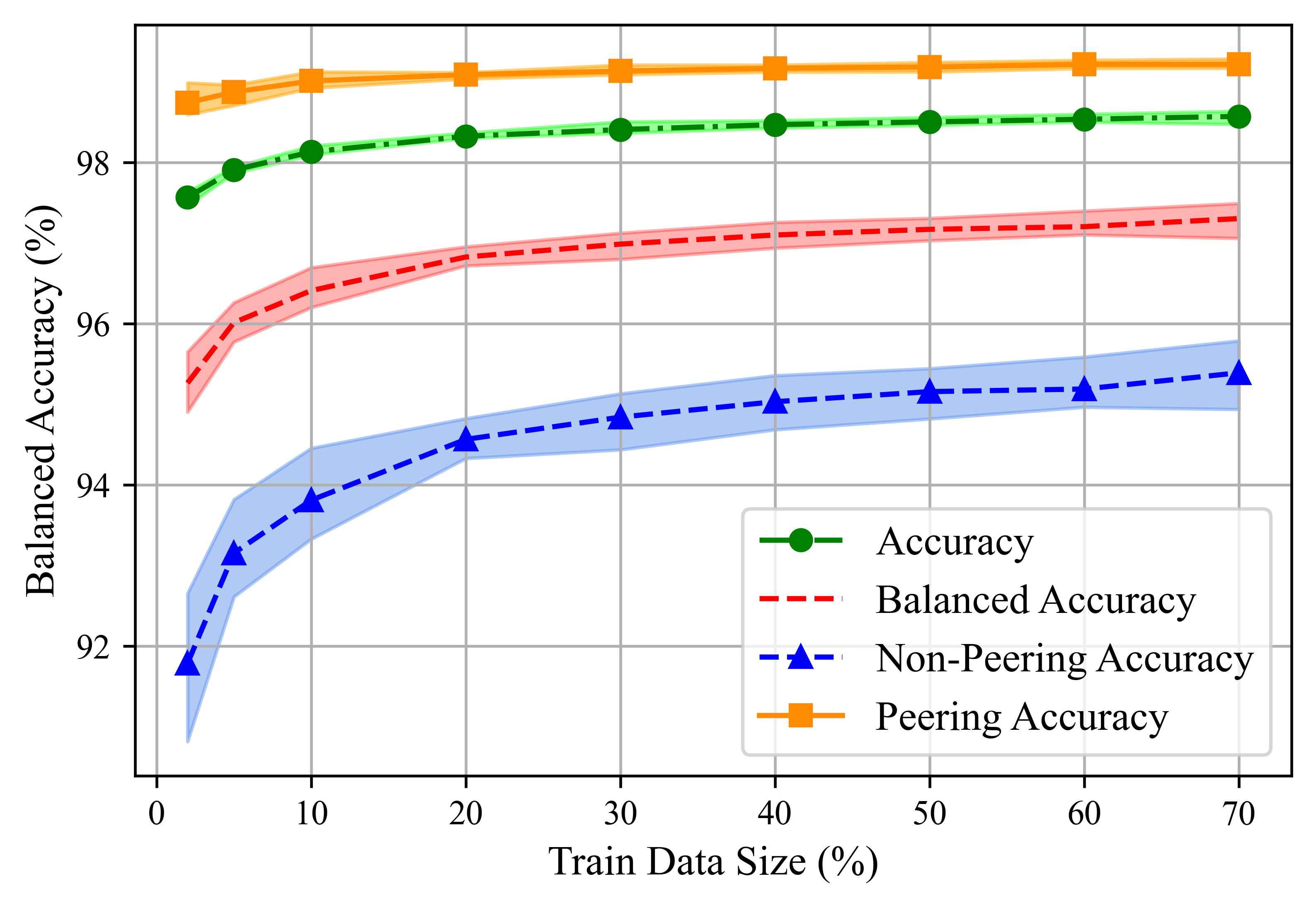}
    \caption{Performance comparison with training data size.}
    \label{fig:perf_train_data}
\end{figure}

The performance of any ML model depends on the data size and how well the ML model can map that data to intelligently generate the required output. Fig. \ref{fig:perf_train_data} shows the performance of the ML model (XGB) in predicting the peering partner when the training data size has been varied from $2\%$ to $70\%$ of the full data size. As expected the model performed well with more training data. However, the most notable part of the figure is that even with only $2\%$ data, i.e., around $7500$ examples of AS pair features and corresponding peering status, the ML model was able to correctly predict if two ASes should peer or not with an overall accuracy of $97.5\%$ (balanced accuracy of $95.2\%$). Moreover, the error interval (the shaded regions) shows that different ML model seeds and training and test data split do not affect the performance much. Hence, the ML model is quite robust to changes in data distribution.

\subsection{Transfer Learning}

\begin{figure}[t]
    \centering
    \includegraphics[width=\linewidth]{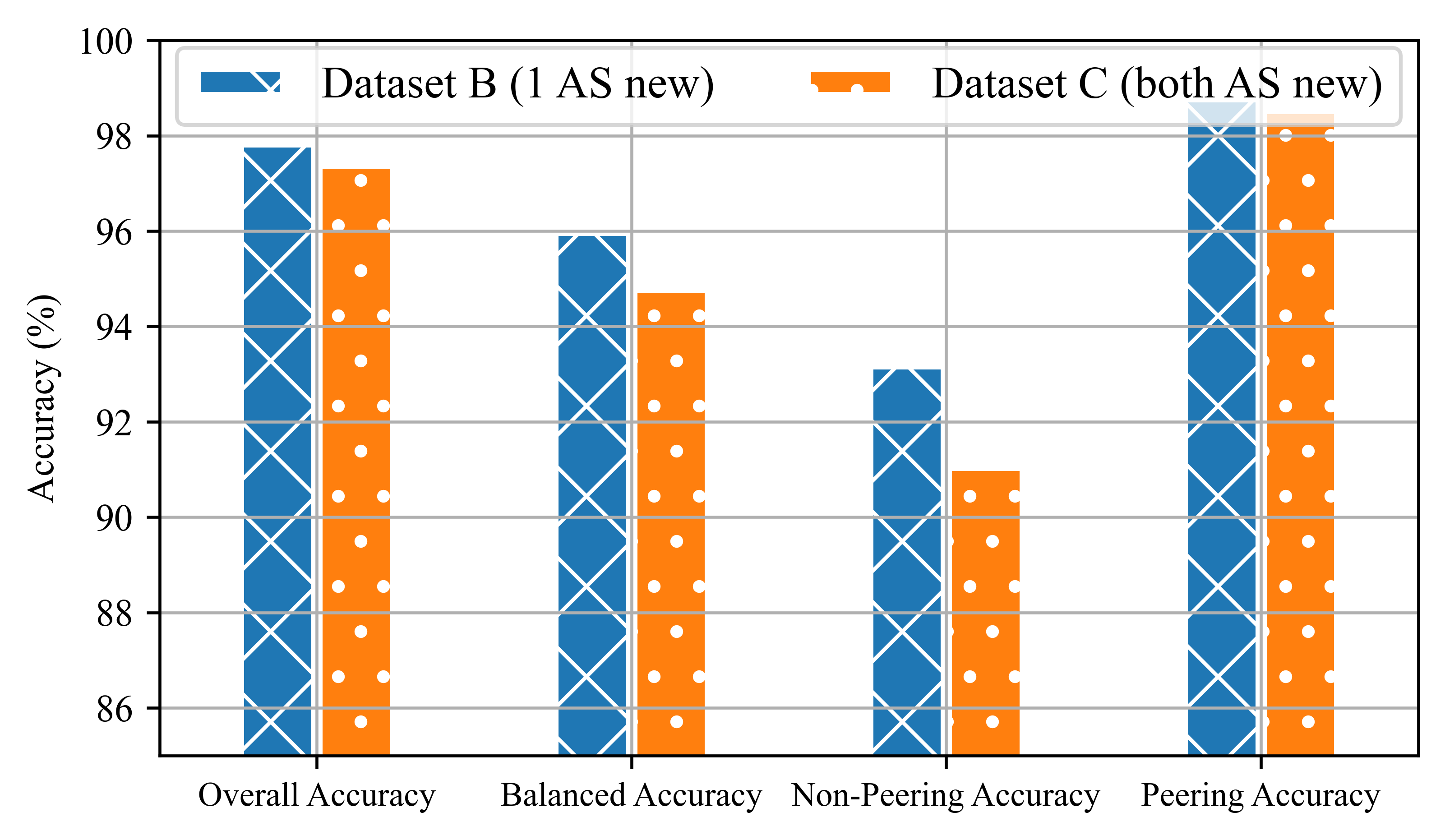}
    \caption{Performance on new AS pairs - transfer learning (same timeline).}
    \label{fig:perf_transfer_learning}
\end{figure}

Up to this point, all the results have been generated with a random train-test split of the whole dataset. However, there is a small problem with that method. With a random train-test split, the training data is indirectly sharing attributes of the ASes present in the test dataset, and that may be a reason for better performance. Hence, we have formulated a technique that minimizes this indirect pollution. We divide the whole ASes into two disjoint sets ($50:50$ ratio), let us call them Set $1$ and Set $2$. Then, we generate three datasets ($A, B$ and $C$) having AS pairs status and corresponding features that obey the following rules: i) The ASes in the AS pairs of Dataset $A$ belongs to Set $1$ only, ii) For Dataset $B$, one of the AS in an AS pair belongs to Set $1$, while the other belongs to Set $2$, and iii) The ASes in AS pairs of Dataset $C$ belongs to Set $2$ only. Now we can use Dataset $A$ as the training dataset and use Dataset $B$ and $C$ as two test datasets. The use of dataset $B$ as a test dataset is important from the perspective of real life as it inherently portrays the case when some new AS has been introduced to the market and we need to know if it should peer with the other ASes already operating in the market. Moving forward, dataset $C$ extends the idea to one step forward. Dataset $C$ symbolizes the introduction of all new ASes to the market and asking if they should peer within themselves. Hence, from the perspective of real-world scenarios, these two datasets are crucial. Fig. \ref{fig:perf_transfer_learning} shows the performance of the ML model when trained on Dataset $A$ and tested on Dataset $B$ and $C$. Interestingly, the ML model performed quite well even on Dataset $B$ and $C$. It was expected that dataset $C$ would not perform as well as dataset $B$, however, a balanced accuracy of $94.5\%$ and overall accuracy of $97.2\%$ indicates that our model can predict peering relationships of totally new ASes with quite high precision.

\begin{remarks}
\newtext{Dataset $B$ portrays the case when some new AS has been introduced to the market while Dataset $C$ symbolizes the introduction of all new ASes to the market.}
\end{remarks}

\subsection{Transfer Learning with time variance}

\begin{figure}[t]
    \centering
    \includegraphics[width=\linewidth]{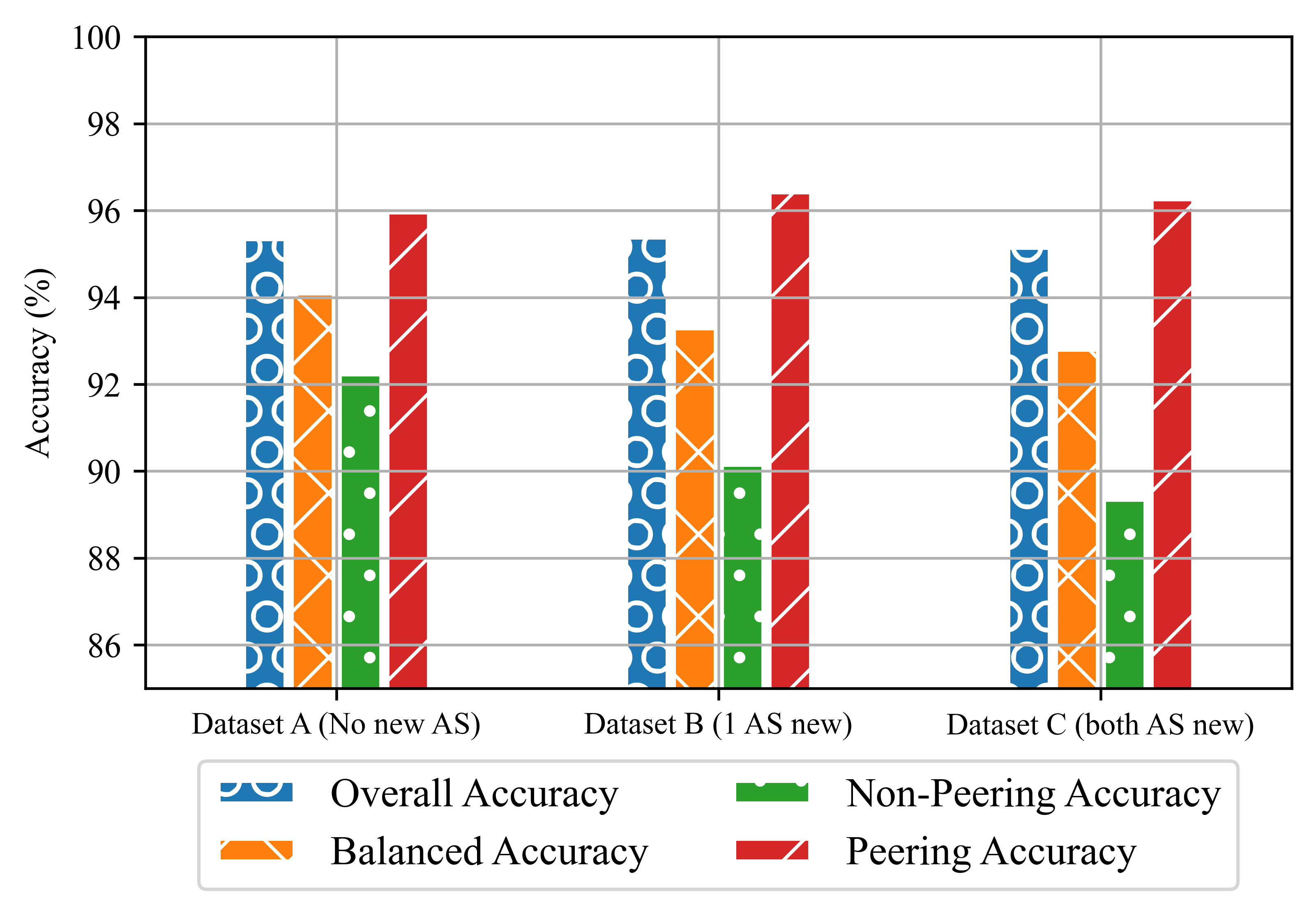}
    \caption{Performance on new AS pairs - transfer learning (different timeline).}
    \label{fig:perf_time_variance}
\end{figure}

To extend the robustness study, let us train the model with data collected at an earlier date (June 2022), and check the performance on data collected later (June 2024). To do so, we use Dataset $A, B$, and $C$ as described before (generated using June 2024 data), but introduce another dataset for training purposes and call that Dataset $O$. In our current settings Datasets $A, B$, and $C$ are generated using data from 2024 (PeeringDB net, AS rank, and AS relationship data). To generate Dataset $O$, we used data from 2022, i.e., the PeeringDB net, AS rank, and AS relationship data from 2022 are used. In Dataset $O$, we have used those AS pairs that are also present in Dataset $A$. Hence, Dataset $O$ is an older version of Dataset $A$, with Dataset $A$ having updated features and peering status of the AS pairs. After the generation of Dataset $O$, we trained the ML model (XGB) with Dataset $O$ and tested on Datasets $A, B$, and $C$. Interestingly, even with the time difference of two years, the trained model performed quite well in all three datasets. As expected, we see the best performance of the ML model on Dataset $A$, and the worst for Dataset $C$. However, even for Dataset $C$, the results are quite good with $95\%$ accuracy. \newtext{Lastly, although the results show that the ML model remains relatively robust to data variance over time, the availability of frequently updated data sources (e.g., PeeringDB and CAIDA) and the extremely fast training time of the XGB model (approximately 15 seconds with the full dataset) suggest that the best practice may be to retrain the model every few days to ensure optimal performance.}

\begin{remarks}
    \newtext{The performance of the ML model is quite robust over time, however due to the fast training time (around 15 seconds), training the model every few days might be the optimal practice.}
\end{remarks}

\subsection{Effect of missing data}

\begin{figure}[t]
    \centering
    \includegraphics[width=\linewidth]{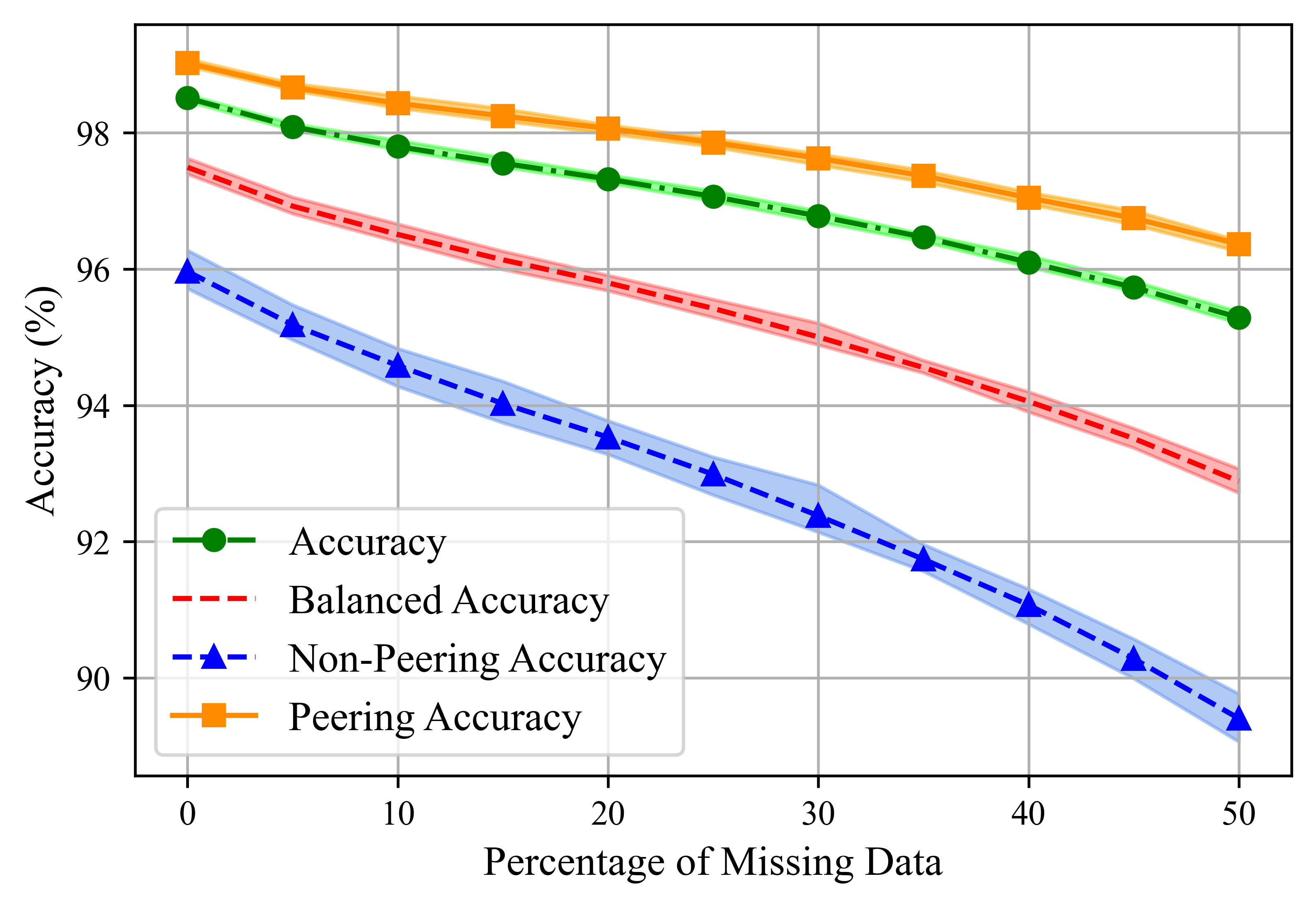}
    \caption{Performance comparison with varying missing data.}
    \label{fig:perf_data_missing}
\end{figure}

To conclude our robustness study, we evaluate the effect of missing values in the data. Say that we want to create a dataset with $10\%$ missing data, then for each feature (column) in the dataset, we randomly delete $10\%$ values from that feature column. Fig. \ref{fig:perf_data_missing} shows the performance with the percentage variation of missing data \footnote{The performance testing for this specific case was done with random train-test split.}. We observe that although missing data decreases the performance, even with $50\%$ of missing data, the model can still predict with $93\%$ balanced accuracy and $95.5\%$ overall accuracy. 

% \subsection{Effect of similar type of ISPs} If all training test data only have ISPs that are content-content. – 1 figure

% \subsection{ISP id ratio to size (IP) ratio – correlation} If closely correlated that explains why ISP id has high feature importance.

\section{Conclusion}
\label{sec:conclusion}

Peering between Internet Service Providers (ISPs) facilitates global connectivity for ISPs. Automation of the peering process can help the ISPs to save money and time, and eventually help to attain efficient Internet connection. In this work, we have explored the use of publicly available data to predict the peering relationship between ISP pairs. CAIDA and PeeringDB maintain two public repositories that have data on ISPs. Moreover, the peering relationship status between ISP pairs is available in the CAIDA AS relationship dataset. Using the feature data on ISPs from CAIDA and PeeringDB, we can create a feature set for each ISP pair that can be used as the input feature for a machine learning model. The peering status of ISPs can be used as the output to train the machine learning model to predict the peering status of ISPs given the feature set of those ISP pairs. Our investigation shows that XGBoost is the best ML algorithm for the peering partner prediction task and can achieve high accuracy. Although PeeringDB features are important in predicting peering partners, CAIDA features were the most relevant. Upon detailed performance analysis, we found the optimum feature set that gives the best performance when used as an input to the ML model. Using the optimal data set, which was formed with the optimal feature sets of the ISP pairs, we achieved more than 98\% accuracy on the peering partner prediction. A robustness study on the model performance shows that the model can learn to predict peering partners quickly. With training data that is only 2\% of the entire dataset, the model can have a prediction accuracy of more than 97\%. In addition, it was observed that the model performance is quite robust to the introduction of new ISPs and consistent with the variation of time. Lastly, the model's performance is also very robust to missing data, and even with 50\% of missing data, the model showed an accuracy of more than 94\%. 
\newtext{Peering partner selection is traditionally done manually, and our results show that a significant part, if not all, of peer selection/prediction can be automated by data-driven peering models powered by ML methods.} We expect that the peering partner prediction process can be fully automated by following our method and ISPs will use the method globally. 
Furthermore, in future work, we aim to explore a data-driven model to solve the peering location problem, i.e., where an ISP pair should peer.

\color{blue}
\section*{APPENDIX}

\begin{figure*}[t]
    \centering
    \includegraphics[width=0.95\linewidth]{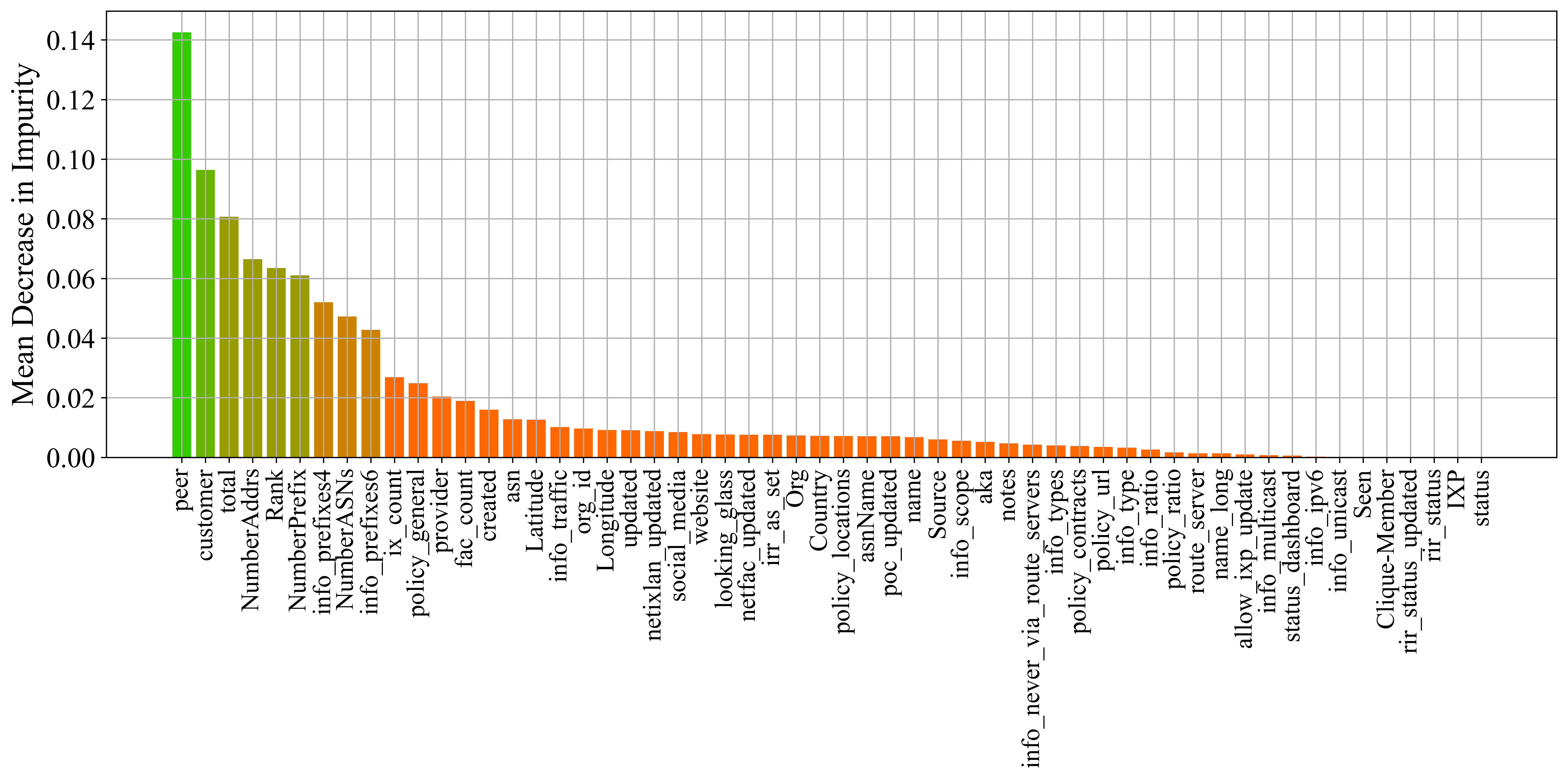}
    \caption{Feature importance of all the features}
    \label{fig:feature_imprt_ALL}
\end{figure*}

\begin{figure*}[t]
    \centering
    \includegraphics[width=0.95\linewidth]{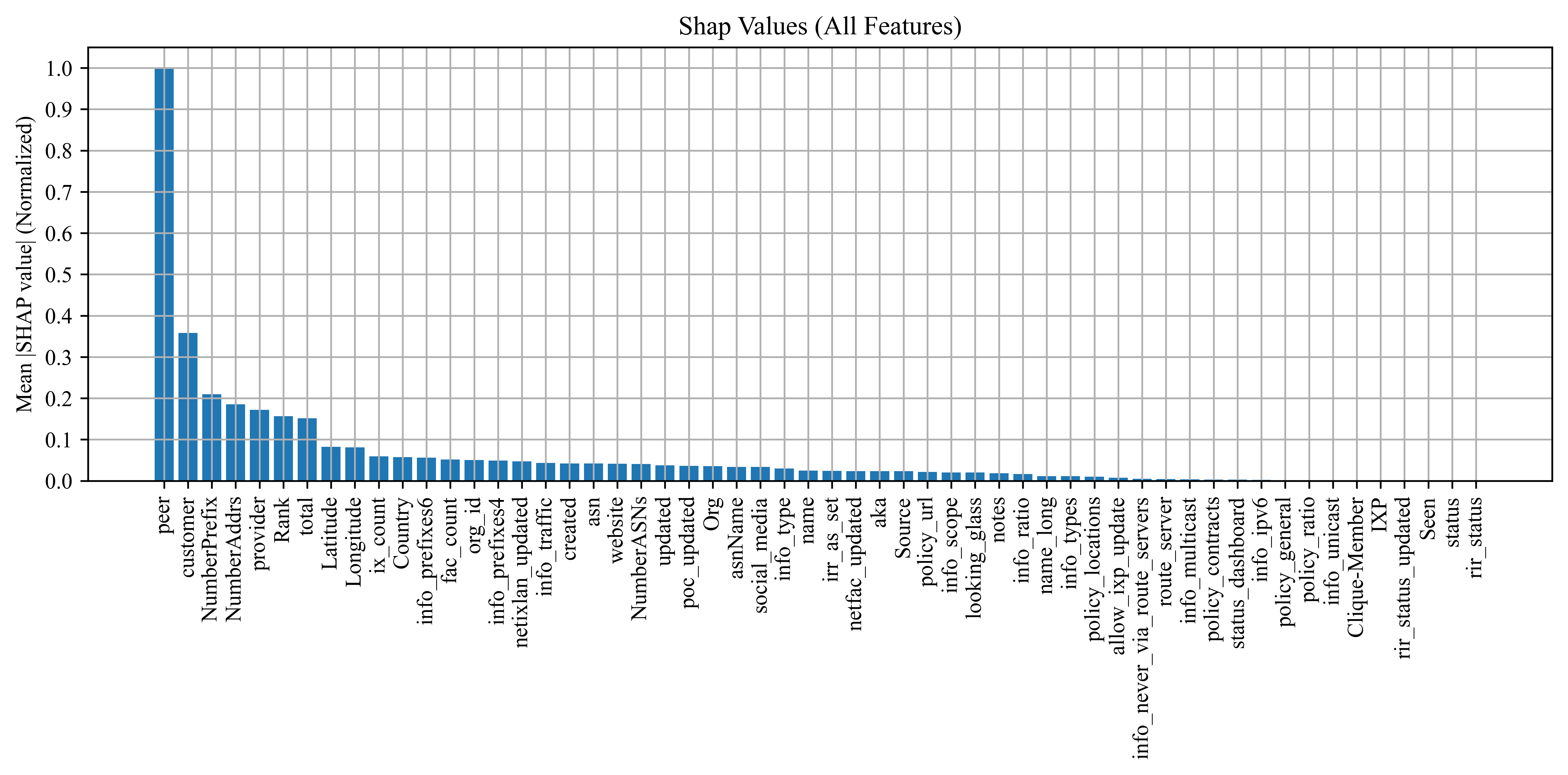}
    \caption{Shapley values of all the features}
    \label{fig:Shap_value_ALL}
\end{figure*}

\subsection{PeeringDB and CAIDA features}
\label{appndx:features}

In CAIDA, there are a total of 18 features available per ISP; the name of these features are:
{\em asn, total, customer, peer, provider, asnName, Clique-Member, NumberASNs, NumberPrefix, NumberAddrs, Country, IXP, Latitude, Longitude, Org, Rank, Seen, Source.}

For PeeringDB, there are a total of 40 features per ISP; the feature names are as follows:
{\em asn, id, status, looking\_glass, route\_server, fac\_count, netixlan\_updated, info\_ratio, policy\_ratio, status\_dashboard, info\_unicast, rir\_status, created, name\_long, policy\_general, website, allow\_ixp\_update, updated, info\_types, rir\_status\_updated, netfac\_updated, info\_traffic, info\_multicast, policy\_locations, name, info\_scope, notes, ix\_count, org\_id, policy\_url, info\_never\_via\_route\_servers, poc\_updated, info\_type, social\_media, policy\_contracts, info\_prefixes6, aka, info\_prefixes4, info\_ipv6, irr\_as\_set.}

\subsection{Feature Removed}
\label{appndx:feature_imprtnc}
There are in total $56$ features per ISP (following the feature extraction procedure in Section \ref{subsubsec:feature_extr}); to evaluate their significance in peering partner prediction, the feature importance scores and the Shapley values were calculated for each of the features.

Fig. \ref{fig:feature_imprt_ALL} depicts the feature importance across all $56$ features (calculated by fitting an RF model with the whole data). Interestingly, many of these features exhibit very low feature importances, with 7 of them having near-zero values. Similarly, Fig. \ref{fig:Shap_value_ALL} presents the Shapley values for all the features. To calculate the Shapley values, we first calculated the Shapley value for each feature across all $373,323$ AS pairs and their corresponding peering labels. The final Shapley value for each feature is calculated by taking the average of the $373,323$ Shapley values of that corresponding feature. Similar to the feature importance values, there are many features with very low Shapley values, with 9 of them having near-zero values, indicating negligible influence on peering prediction.

\begin{figure}[t]
    \centering
 \includegraphics[width=0.95\linewidth]{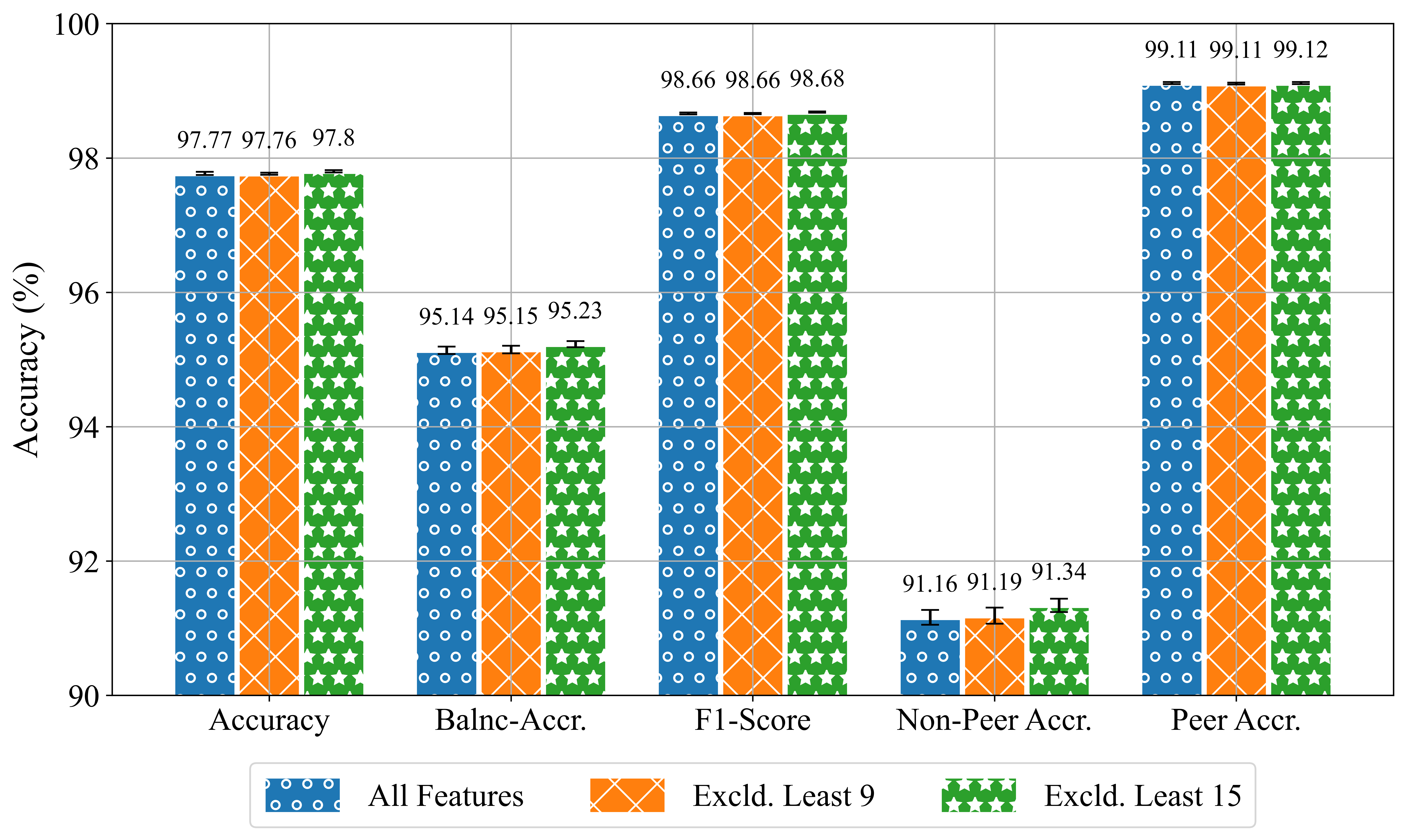}
    \caption{Performance evaluation by removing features having low Shapley values.}
    \label{fig:performance_low_feature_remv}
\end{figure}

From the results of feature importance and Shapley values, we decided to evaluate performance by removing the features that have significantly low Shapley values. 
%(less than 1\% compared to the highest Shapley value)
While doing so, we also checked the feature importance values of the corresponding features to make sure that it is actually insignificant. Fig. \ref{fig:performance_low_feature_remv} shows the performance of the RF based ML model when 9 features having near-zero Shapley values were removed as well as when 6 additional features with very low Shapley values were removed.
%So, in essence, along with the , we removed total 15 features from the full feature set. 
The results show that removing the low Shapley valued features increases the overall performance across all metrics (F1 score, accuracy values, etc.). Hence, we decided to remove the 15 features from the ISP feature set. Those features are:
{\em Clique-Member, IXP, Seen, status, policy\_ratio, info\_unicast, rir\_status, policy\_general, rir\_status\_updated, route\_server, status\_dashboard, info\_multicast,  policy\_contracts, info\_never\_via\_route\_servers, info\_ipv6.}
Among these 15 features, 3 are from CAIDA, and 12 are from PeeringDB. As a result, we are left with $56-15 = 41$ features.

\newtext{Interestingly, among the 15 features that were removed, three were related to peering policy. A possible reason for this could be that many of the values for these features were missing in PeeringDB. To ensure that we were not overlooking important information from the textual data, we also trained a Longformer model using all features, including their textual values. However, we did not observe any evidence that peering policies could meaningfully enhance performance given the current missing and incomplete data. It is possible that if PeeringDB provided comprehensive peering policies for all ISPs, their significance might become apparent, but this is not something we can confirm with the data currently available.}

\subsection{Filtering Features}
\label{appndx:drop_featrs_1by1}

\begin{figure}[t]
    \centering
    \includegraphics[width=\linewidth]{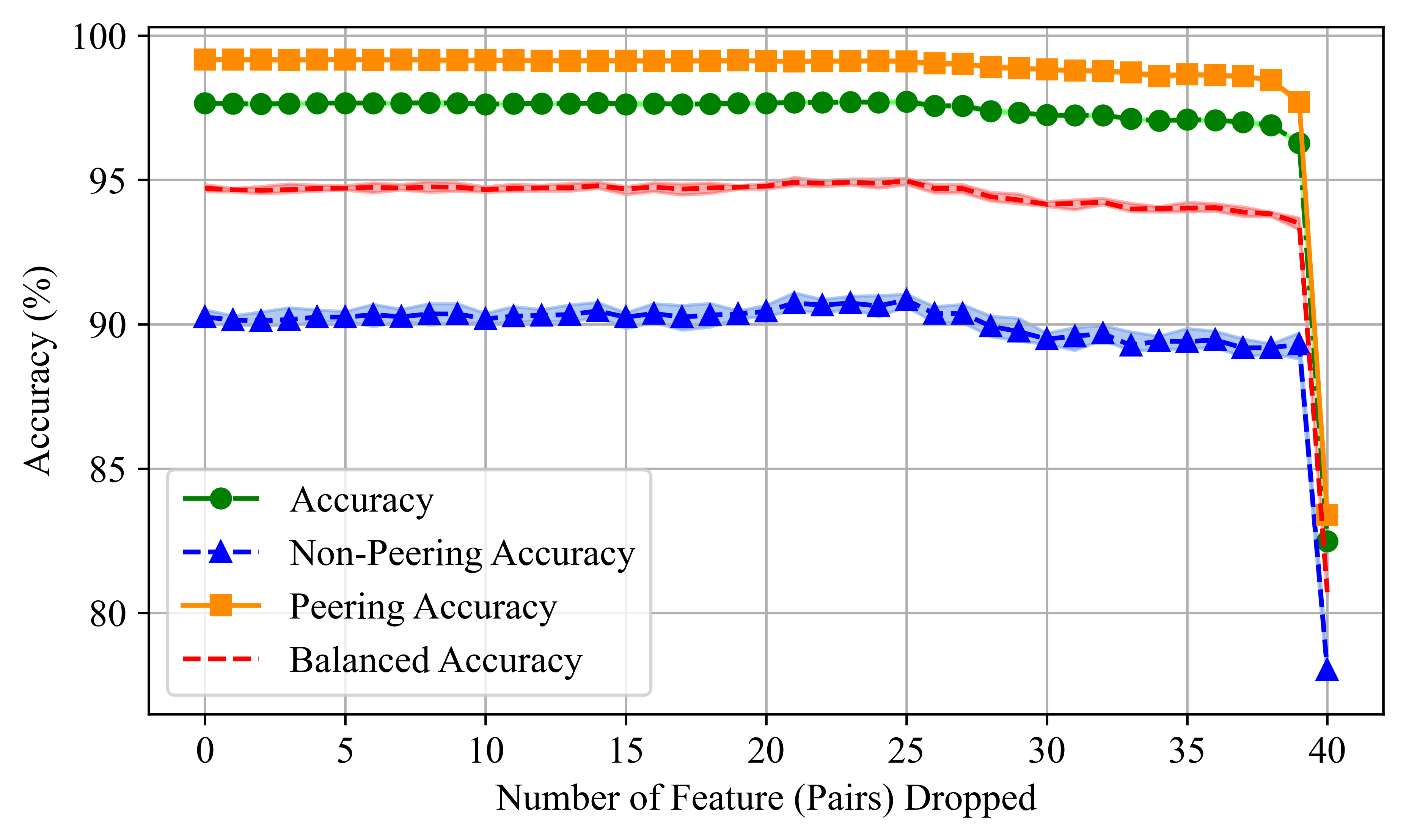}
    \caption{Accuracy of RF model by sequentially dropping least important feature (pairs).}
    \label{fig:FI_drop_accr}
\end{figure}

The performance of the RF based ML model when the least important features were dropped one pair at a time is given in Fig. \ref{fig:FI_drop_accr}. Since 15 features were dropped previously from the original 56 features, the current result reflects the performance of the top 41 remaining features. The list of features in terms of their least importance (when dropping one by one) is as follows:
{\em 1) allow\_ixp\_update,
2) name\_long,
3) info\_type,
4) info\_ratio,
5) notes,
6) info\_types,
7) policy\_url,
8) info\_scope,
9) Source,
10) policy\_locations,
11) aka,
12) asnName,
13) looking\_glass,
14) name,
15) Country,
16) website,
17) poc\_updated,
18) irr\_as\_set,
19) Org,
20) info\_traffic,
21) netfac\_updated,
22) netixlan\_updated,
23) social\_media,
24) updated,
25) org\_id,
26) Longitude,
27) created,
28) Latitude,
29) asn,
30) provider,
31) fac\_count,
32) info\_prefixes6,
33) ix\_count,
34) info\_prefixes4,
35) NumberASNs,
36) Rank,
37) total,
38) NumberPrefix,
39) NumberAddrs,
40) peer,
41) customer.}

As shown in the figure, the ML model performance remains almost constant when the 20 least important features are dropped. There is a slight improvement when up to 25 features are dropped, with peak performance occurring around the removal of the 24th-25th least important feature, which is more clearly visualized in Fig. \ref{fig:FI_drop_accr}. Dropping more features beyond the 25th feature showed a steady decrease in performance. Finally, even with only the two most important features, `$peer$' and `$customer$', the ML model achieves an accuracy of 96.5\% and a balanced accuracy of 93.5\%. Lastly, with only the single most important feature, `$customer$', the ML model's performance drops to an accuracy of 88\% and to a balanced accuracy of 86\%.

\color{black}

% \section*{REFERENCES}

\bibliographystyle{IEEEtran}
\bibliography{bib_file/main}

\newpage

\pagebreak

\newpage

\vfill\pagebreak

\end{document}